\pdfoutput=1

\documentclass[11pt]{article}

\usepackage[]{EMNLP2023}

\usepackage{times}
\usepackage{latexsym}
\usepackage{multirow}
\usepackage[T1]{fontenc}

\usepackage[utf8]{inputenc}

\usepackage{microtype}

\usepackage{inconsolata}

\usepackage{multirow}
\usepackage{graphicx}
\usepackage{booktabs}
\usepackage{amssymb}

\usepackage{microtype}

\usepackage{soul}
\usepackage{pifont}
\usepackage{times}
\usepackage{latexsym}
\usepackage{graphicx}
\usepackage{amsmath}
\usepackage{subcaption}

\usepackage{amsmath}
\usepackage{bbold}
\usepackage{mathtools}
\usepackage{xcolor}

\usepackage{tabularx,ragged2e}
\usepackage{enumitem}
\usepackage{scrextend}

\newcommand{\cmark}{\ding{51}}
\newcommand{\xmark}{\ding{55}}

\DeclareRobustCommand{\huggingface}{%
  \begingroup\normalfont
  \vspace{-0.2em}%
  \raisebox{-0.4em}{%
  \includegraphics[height=1.5em]{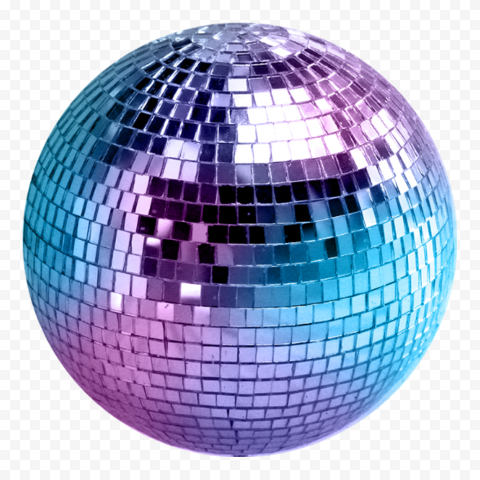}%
  }%
  \kern 0.4em%
  \endgroup
}

\title{EDM3: Event Detection as Multi-task Text Generation}

\author{
Ujjwala Anantheswaran\thanks{~~Now at Microsoft Corporation} \hspace{9pt}Himanshu Gupta\hspace{9pt}Mihir Parmar  \\ \hspace{9pt} \textbf{Kuntal Kumar Pal} \hspace{9pt}
\textbf{Chitta Baral}\\
 Arizona State University\\
 \tt\small   \texttt{\{uananthe, hgupta35, mparmar3, kkpal, chitta\}}@asu.edu
}

\begin{document}

\maketitle
\begin{abstract}
Event detection refers to identifying event occurrences in a text and comprises of two subtasks; event identification and classification.
We present EDM3, a novel approach for Event Detection that formulates three generative tasks: identification, classification, and combined detection. 
We show that EDM3 helps to learn transferable knowledge that can be leveraged to perform Event Detection and its subtasks concurrently, mitigating the error propagation inherent in pipelined approaches. 
Unlike previous dataset- or domain-specific approaches, EDM3 utilizes the existing knowledge of language models, allowing it to be trained over any classification schema.
We evaluate EDM3 on multiple event detection datasets: RAMS, WikiEvents, MAVEN, and MLEE, showing that EDM3 outperforms 1) single-task performance by 8.4\% on average and 2) multi-task performance without instructional prompts by 2.4\% on average.
We obtain SOTA results on RAMS (71.3\% vs. 65.1\% F-1) and competitive performance on other datasets.
We analyze our approach to demonstrate its efficacy in low-resource and multi-sentence settings. 
We also show the effectiveness of this approach on non-standard event configurations such as multi-word and multi-class event triggers. 
Overall, our results show that EDM3 is a promising approach for Event Detection that has the potential for real-world applications
\footnote{Data and source code are available at \url{https://github.com/ujjwalaananth/EDM3_EventDetection}}.
\end{abstract}

\section{Introduction}

Event Detection (ED) is a fundamental task in natural language processing that involves identifying the occurrence and intent of an event from unstructured text, by recognizing its \textit{event triggers} and assigning it to an appropriate \textit{event type}. 
The event type is defined by a schema that characterizes the event's nature and specifies the scope of roles involved in understanding the event. 
ED has a wide range of applications in various downstream tasks, such as information retrieval \cite{10.1145/2911451.2914805}, event prediction \cite{souza2020event}, and implicit argument detection \cite{cheng-erk-2018-implicit}. 
Typically, ED comprises two subtasks: Event Identification (EI), which is the identification of an event trigger, and Event Classification (EC), or the classification of the identified trigger. 


Existing methods for ED cannot easily leverage pretrained semantic knowledge \cite{lai2020event}.
These models fall short of correctly identifying complex events and face difficulties in few-shot ED settings. 
Lastly, these models, once trained, lack cross-domain or cross-task adaptability. 
The subpar performance of these event detection modules may handicap the overall efficacy of pipelined event extraction systems \cite{liu-etal-2020-event}. 

We address the above challenges by proposing a new training paradigm wherein we train a generative model on ED alongside its constituent subtasks, called EDM3 i.e. \textbf{E}vent \textbf{D}etection by \textbf{M}ulti-task Text Generation over \textbf{3} subtasks. 
We show that by modeling ED and its subtasks as individual, similarly-formatted sequence generation tasks, a model can learn transferable knowledge from the subtasks that can be leveraged to improve performance on ED. 
In contrast with the conventional token classification discriminative approaches, EDM3 leverage text-to-text generation methods which give the advantage to perform individual subtasks in a non-pipelined fashion.
To the best of our knowledge, this work is the first to utilize all ED subtasks separately and jointly, while moving away from the traditional token classification paradigm. 
EDM3 also generalizes well without the need for the creation of domain-specific embeddings. 
Table \ref{teaser_tab} highlights the advantages of EDM3 over previous SOTA approaches.

\begin{table*}[]
\resizebox{16cm}{!}{%
\begin{tabular}{ccccccc}
\toprule
\multicolumn{1}{c}{\multirow{2}{*}{\textbf{Approaches}}} &
  \multirow{2}{*}{\textbf{\begin{tabular}[c]{@{}c@{}}Datasets\end{tabular}}} &
  \multicolumn{3}{c}{\textbf{Tasks Covered}} &
  \multirow{2}{*}{\textbf{\begin{tabular}[c]{@{}c@{}}Domain \\ Genralization\end{tabular}}} &
  \multirow{2}{*}{\textbf{\begin{tabular}[c]{@{}c@{}}Comparative \\ Performance\end{tabular}}} \\ \cmidrule{3-5}
\multicolumn{1}{c}{} &
   &
  \multicolumn{1}{l}{\textbf{Identification}} &
  \textbf{Classification} &
  \textbf{Detection} &
   &
   \\ \midrule
\citet{liu-etal-2022-saliency} & ACE, MAVEN           & \xmark                & \xmark & \cmark & \xmark                & SOTA on MAVEN       \\
                       & \multicolumn{1}{l}{} & \multicolumn{1}{l}{} &       &           & \multicolumn{1}{l}{} & \multicolumn{1}{l}{} \\
\citet{Veyseh2021UnleashGP} &
  ACE, RAMS, CysecED &
  \cmark &
  \xmark &
  \cmark &
  \cmark &
  \begin{tabular}[c]{@{}c@{}}SOTA on ACE and CysecED\\ Competitive on RAMS\end{tabular} \\
                       & \multicolumn{1}{l}{} & \multicolumn{1}{l}{} &       &           & \multicolumn{1}{l}{} & \multicolumn{1}{l}{} \\
\citet{2022A}                  & MLEE                 & \xmark                & \xmark & \cmark & \xmark                & SOTA on MLEE         \\ \midrule
\begin{tabular}[c]{@{}l@{}}EDM3 (Ours)\end{tabular} &
  \begin{tabular}[c]{@{}c@{}}MAVEN, MLEE\\ WikiEvents, RAMS\end{tabular} &
  \cmark &
  \cmark &
  \cmark &
  \cmark &
  \begin{tabular}[c]{@{}c@{}}SOTA on RAMS\\ Competitive on MLEE \& MAVEN\\ Benchmark on WikiEvents\end{tabular} \\ \bottomrule
\end{tabular}
}
\caption{Comparison of EDM3 with other state-of-the-art approaches highlighting the advantages of the approach over them. 
Columns `Identification', `Classification', and `Detection' denote whether the approach can be used to perform these tasks independently and end-to-end with no model modification. ``Domain Genralization'' refers to the ability to perform ED or its subtasks over non-general domains as well as on the general domain. We add some additional information to provide the context in terms of performance metrics. 
EDM3 demonstrates high efficacy on multiple domains and can be leveraged to perform identification, classification, and combined detection, independently and in an end-to-end fashion without model modification, over all the domains it is used on. }
\label{teaser_tab}
\end{table*}

We conduct extensive experiments using T5-base model \cite{raffel2020exploring} on RAMS, WikiEvents, MAVEN, and MLEE datasets. 
We achieve an $F_1$ score of 71.3\% on RAMS, surpassing GPTEDOT's \cite{Veyseh2021UnleashGP} score of 65.1\%. 
This training approach also helps achieve competitive performance on the MAVEN dataset, obtaining 58.0\% $F_1$.
We also fine-tune the same domain-agnostic model on the biomedical-domain MLEE dataset to obtain a score of 78.1\%.
Finally, we also establish the benchmark performance on the ED task for the WikiEvents dataset with 60.7\% $F_1$ score. 
The results on the aforementioned datasets not only demonstrate the paradigm's cross-domain adaptability but also show its efficiency, as a T5-base (220M) model achieves a SOTA and competitive score compared to more sophisticated or domain-specific modeling approaches.

We conduct investigations along multiple lines of inquiry to get notable insights. 
This includes an exploration of the efficacy of our approach in adapting to low-resource event scenarios, the effects of multi-tasking, and using instructional prompts, with a further dive into the type of instructional prompts that add the most value. 
We also evaluate the performance of this approach over non-standard event configurations, such as multi-word and multi-class triggers, which are notably absent from benchmark datasets but are highly prevalent in real-world data. 
Finally, we discuss the influence of the data on the effectiveness of the paradigm, focusing on the presence of negative examples and the importance of contextual information.
In summary, our contributions are as follows:
\begin{enumerate}[noitemsep,nosep,leftmargin=*]

    \item We propose ED as a division of subtasks converted into \textit{sequence generation (text-to-text)} format which utilizes transferable knowledge from atomic tasks to improve performance on a complex primary task (ED).
    \item We use this unified paradigm to obtain SOTA or competitive performances over various datasets across multiple domains.
    \item We perform a methodical analysis of the impact of our method on various facets of the task and demonstrate its efficacy over complex real-world scenarios.
\end{enumerate}

\section{Related Work}

Transformer-based models \cite{NIPS2017_3f5ee243} have been at the forefront of many language tasks due to the wealth of pretrained knowledge. 
Models using BERT \cite{yang-etal-2019-exploring,wang-etal-2019-adversarial-training} treat ED as word classification, in graph-based architectures \cite{wadden-etal-2019-entity,lin-etal-2020-joint}. 
Models that improve ED performance for low resource settings include \citet{lu-etal-2019-distilling,deng-etal-2021-ontoed}. 
Other works \cite{tong-etal-2020-improving,Veyseh2021UnleashGP} generate ED and EI samples respectively to augment training data. 
Many models frame ED as a question-answering task \cite{du-cardie-2020-event,DBLP:journals/corr/abs-2104-06969,DBLP:journals/corr/abs-2110-07476,liu-etal-2020-event}. 
APEX \citep{wang2022art} augments input with type-specific prompts. With the advent of more powerful sequence-to-sequence models such as T5, there has been an increased interest in formulating event detection and event extraction as sequence generation tasks \cite{tanl, lu-etal-2021-text2event, si2022generating}\footnote{Extended related work is discussed in App. \ref{sec:Other_Related_Work}}. 

\begin{figure*}[t!]
	\centering
	\includegraphics[width= 0.8\linewidth, height= 3.5cm]{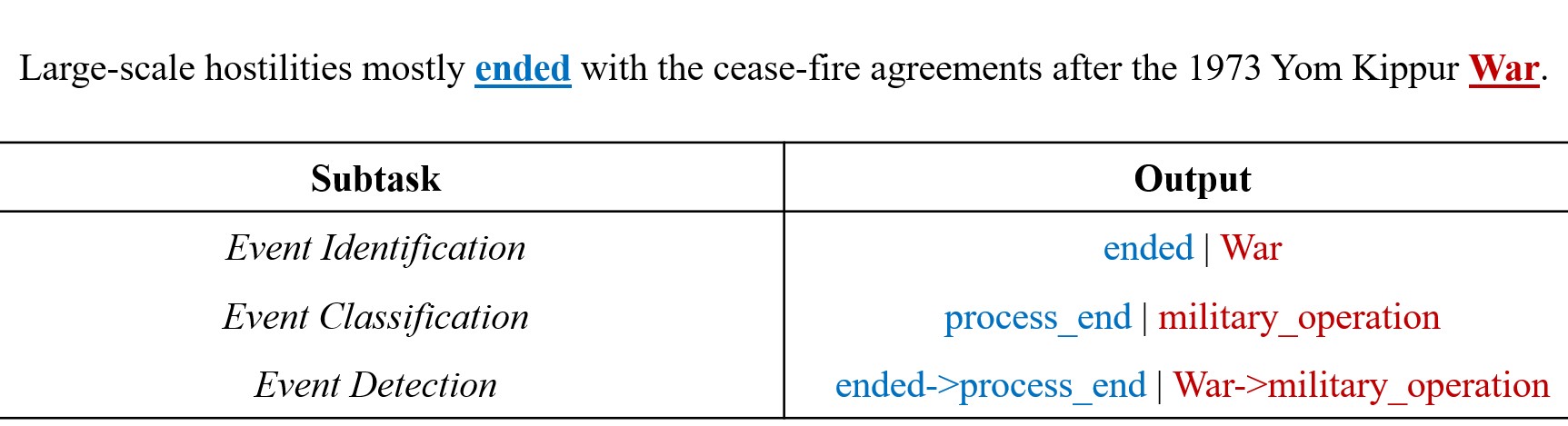}
	\caption{Illustration of generatively reformulated outputs for ED and its subtasks. The outputs for EI and EC are singly-delimited strings containing extracted triggers or event types present in the instance. The output for ED is a doubly-delimited string containing all event triggers and their corresponding event types.}
	\label{fig:3subtasks}
\end{figure*}

\paragraph{Multi-Task Learning} is a training paradigm in which a single machine learning model is trained on multiple separate tasks \citep{10.1023/A:1007379606734,https://doi.org/10.48550/arxiv.2009.09796}. 
Across domains, models trained on multiple disparate tasks are better performing due to shared learning. 
Multi-Task learning has been leveraged to great effect in \citet{xie2022unifiedskg,lourie2021unicorn}, and in specific domains as well \citep{article,parmar2022inboxbart}. 
This paradigm is also the basis of the generative T5 model. 
\citet{tanl} carried out multi-task learning experiments over a number of information retrieval tasks. 
Specifically for Event Detection, multi-tasking over ED subtasks is implemented in GPTEDOT \cite{Veyseh2021UnleashGP}, where EI is used to augment ED performance. 
This is because the simplicity of EI makes it easier to evaluate the quality of generated data. 
However, there is a risk of introducing noise or generating low-quality samples due to the characteristics of the source data.

\paragraph{Prompt engineering} Prompt-based models have been used for Event Detection and Event Extraction as well. 
More recently, \citet{si2022generating} used predicted labels from earlier in the pipeline as prompts for later stages of trigger identification and argument extraction, while \citet{wang2022art}, following the example of other works that use prototype event triggers \citep{wang-cohen-2009-character,bronstein-etal-2015-seed,DBLP:journals/corr/abs-1910-11368,lyu2021zero,liu-etal-2020-event,zhang-etal-2021-zero} from the dataset, used triggers as part of tailored prompts for each event type in the schema. 
In proposing EDM3, we are the first to explore the efficacy of instructional prompts for ED.




\section{Methodology}

Given an input instance containing event triggers of various event types, we aim to identify all the triggers present and classify them. 
As a preliminary step, we decompose and reformulate ED and its subtasks as sequence generation tasks.
Having done so, we train a T5 model on all 3 generative tasks simultaneously to create a single multi-task model. 
We also provide task-specific natural language instructional prompts with illustrative examples. 
Finally, we use beam search decoding to select tokens during sequence generation. 
We delineate these steps in more detail below.

\paragraph{Task Decomposition}
ED is a multi-level task requiring both event identification and classification, which traditional sequence labeling approaches conduct in a single step. 
We decompose ED into independent atomic sequence generation tasks that are carried out in parallel with each other or with the primary task, to augment the training process.

\subsection{Generative reformulation}

The task labels, whether event triggers, event types, or a more comprehensive list of event and corresponding type annotations, are converted to a delimited string. 
This creates a consistent pattern that can be learned by the model.  
In the absence of any events, we use the label NONE.
Due to the presence of multi-class triggers, the number of unique event types and unique triggers for an instance might differ, making all tasks notably distinct from one another, as opposed to ED being simply a linear combination of EI and EC.

\paragraph{Event Identification/Classification}
Each label for these tasks contains a single component, i.e., either the event trigger or the event type. 
Hence, we can represent the output of each instance as a singly-delimited sequence of labels. 
For example, an instance with $x$ unique triggers would have the following label representation for the EI task:
\centerline{\textbf{$T_{1}$} | \textbf{$T_{2}$} | \textbf{$T_{3}$} ... \textbf{$T_{x}$}}
Where $T_{i}$ is the $i^{th}$ event trigger occurring in an input instance. 
Similarly, an instance with $y$ unique event types occurring in it would have the following output representation for the EC task:
\centerline{\textbf{$E_{1}$} | \textbf{$E_{2}$} | \textbf{$E_{3}$} ... \textbf{$E_{y}$}}
Where $E_{i}$ is the $i^{th}$ type of event occurring in the instance. We delimit all triggers and types with a pipe ($|$) symbol.

\paragraph{Event Detection}
Each label for ED is composed of 2 components: the event trigger, and its corresponding event type. Similar to our sequence formulation for EI and EC, we create a doubly-delimited sequence of events for an instance with $x$ events.
We use -> as a delimiter between trigger and type, creating a unique format to enumerate the list of events. This allows us to represent multiple events in an instance as follows:
\centerline{\textbf{$T_{1}$->$E_{1}$} | \textbf{$T_{2}$->$E_{2}$} | \textbf{$T_{3}$->$E_{3}$} ... \textbf{$T_{x}$->$E_{x}$}}

For an example of an instance showing the reformulated outputs for all tasks, see Figure \ref{fig:3subtasks}. 

\subsection{Multi-Task Learning}
We posit that by explicitly modeling individual subtasks concurrently with Event Detection (ED), a multi-task learning model can acquire knowledge that is transferable across atomic tasks, thereby enhancing ED performance. 
Our proposed method involves modeling Entity Classification (EC) separately for rarer event types. 
By doing so, the model gains the ability to explicitly identify instances containing these events, leading to improved identification and classification of their triggers. 



\begin{figure}[t!]
	\centering
	\includegraphics[width= \linewidth, height= 11.5 cm]{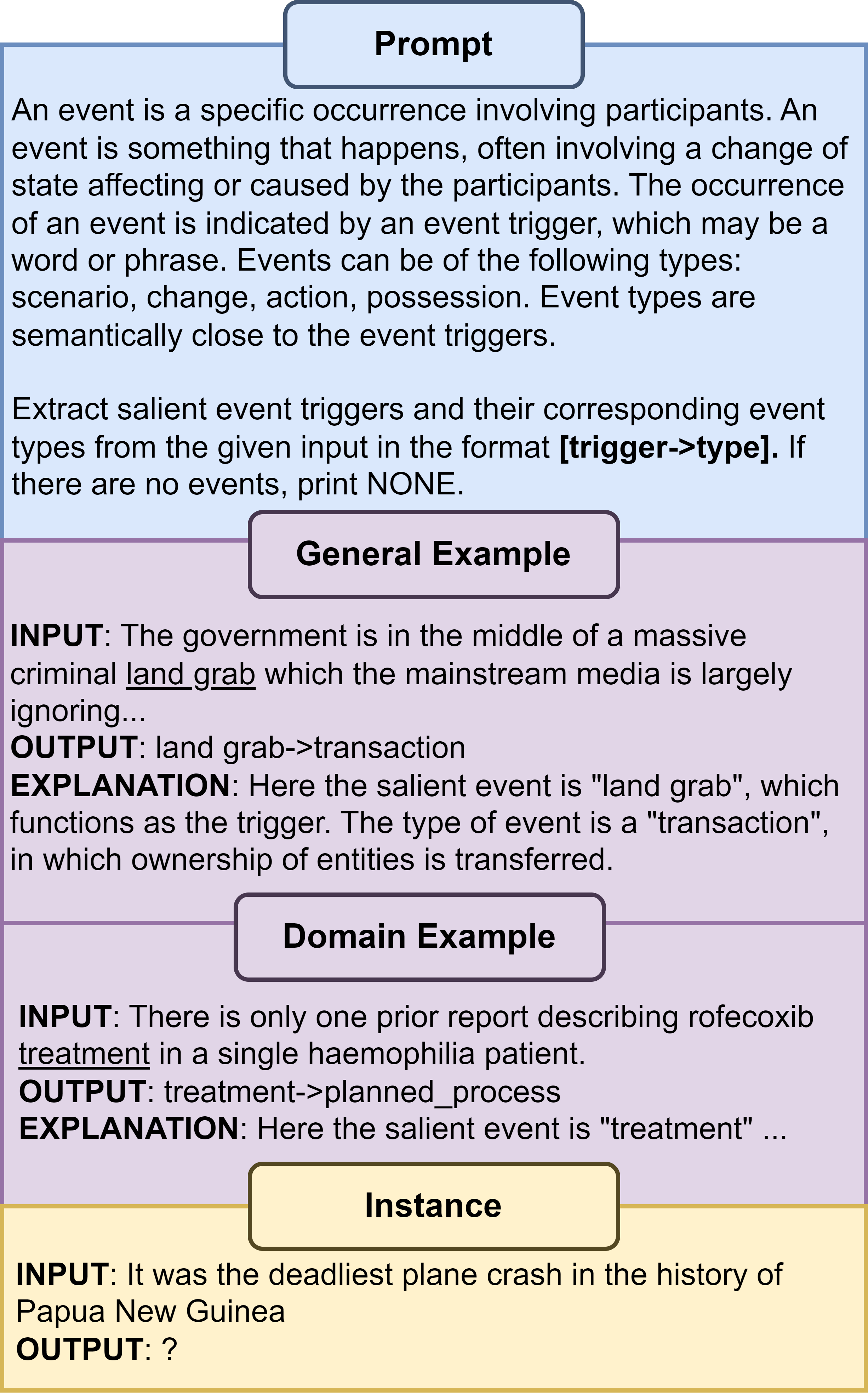}
	\caption{An example of an input instance for reformulated generative ED. The input comprises a task definition followed by diverse domain examples before the input sentence containing the events to be detected.}
	\label{teaser2}
\end{figure} 

\subsection{Instructional Prompt Tuning for Generative ED}
We use instructional prompts to improve multitasking. 
We design natural language prompts that describe how to perform event identification, classification, or detection. 
An illustration of this is mentioned in Figure \ref{teaser2}.

\section{Data}

The datasets we choose to demonstrate our approach on span a range of characteristics, from sentence-level to multi-sentence level, with varying proportions of non-event instances. 
We also include a biomedical domain dataset to illustrate the adaptability of our approach. 
In Table \ref{tab:stats_overview}, we note the document and event instance statistics across datasets. 
Table \ref{tab:stats_proc} delineates the dataset statistics post-data processing. 
We note the average and maximum number of events and distinct event types that occur per data instance for each dataset.

\paragraph{MAVEN} \citet{wang-etal-2020-maven} proposed this dataset with the idea of combating data scarcity and low coverage problem in prevailing general domain event detection datasets. 
The high event coverage provided by MAVEN results in more events per sentence on average, including multi-word triggers, as compared to other general domain ED datasets (more details in App. \ref{app:annotation}). The dataset, reflective of real-world data, has a long tail distribution (see Figure \ref{fig:maven_dtb}). 
We follow the example of SaliencyED {\citep{liu-etal-2022-saliency}} and evaluate our model performance on the development split of the original MAVEN dataset.

\paragraph{WikiEvents} Existing work on this dataset proposed by \citet{Li2021DocumentLevelEA} focuses exclusively on document-level argument extraction and event extraction. 
Sentences without any event occurrences make up nearly half of the entire dataset (see Table \ref{tab:stats_proc}). 
In the absence of existing baselines, we establish the benchmark performances on sentence-level ED on this dataset for future researchers.

\begin{table}[t!]
\centering
\resizebox{\columnwidth}{!}{%
\begin{tabular}{l|ccc|cc}
\toprule
\multirow{2}{*}{\textbf{Dataset}} & \multicolumn{3}{c|}{\textbf{Docs}}                                    & \multicolumn{1}{c}{\multirow{2}{*}{\textbf{\#triggers}}} & \multicolumn{1}{c}{\multirow{2}{*}{\textbf{\#types}}} \\ \cline{2-4}
                         & \multicolumn{1}{l}{\textbf{Train}} & \multicolumn{1}{l}{\textbf{Dev}} & \textbf{Test} & \multicolumn{1}{c}{}                            & \multicolumn{1}{c}{}                         \\ \hline
MLEE                     & \multicolumn{1}{l|}{131}   & \multicolumn{1}{l|}{44}  & 87   & 8014                                             & 30                                            \\ 
RAMS                     & \multicolumn{1}{l|}{3194}  & \multicolumn{1}{l|}{399} & 400  & 9124                                             & 38                                            \\ 
MAVEN                    & \multicolumn{1}{l|}{2913}  & \multicolumn{1}{l|}{710} & 857  & 118732                                           & 168                                           \\ 
WikiEvents               & \multicolumn{1}{l|}{206}   & \multicolumn{1}{l|}{20}  & 20   & 3951                                             & 49                                            \\ \bottomrule
\end{tabular}%
}
\caption{Dataset statistics, including number of documents per data split, as well as number of event triggers and unique event types across the dataset.}
\label{tab:stats_overview}
\end{table}
\begin{table}[t!]
\centering
\resizebox{\columnwidth}{!}{%
\begin{tabular}{l|c|cc|cc|c}
\toprule
\multicolumn{1}{c|}{\multirow{2}{*}{\textbf{Dataset}}} & \multicolumn{1}{c|}{\multirow{2}{*}{\textbf{Neg (\%)}}} & \multicolumn{2}{c|}{\textbf{Events per row}} & \multicolumn{2}{c|}{\textbf{Types per row}} & \multirow{2}{*}{\textbf{\#zs}} \\ \cline{3-6}
\multicolumn{1}{c|}{} & \multicolumn{1}{c|}{} & \multicolumn{1}{c|}{\textbf{Avg}} & \multicolumn{1}{c|}{\textbf{Max}} & \multicolumn{1}{c|}{\textbf{Avg}} & \multicolumn{1}{c|}{\textbf{Max}} &  \\ \hline
MLEE & 18.22 & \multicolumn{1}{l|}{2.867} & 16 & \multicolumn{1}{l|}{2.369} & 9 & 3 \\
RAMS & 0 & \multicolumn{1}{l|}{1.066} & 6 & \multicolumn{1}{l|}{1.061} & 4 & 0 \\
MAVEN & 8.64 & \multicolumn{1}{l|}{2.433} & 15 & \multicolumn{1}{l|}{2.314} & 15 & 0 \\
WikiEvents & 54.11 & \multicolumn{1}{l|}{1.671} & 7 & \multicolumn{1}{l|}{1.429} & 6 & 1 \\ \bottomrule
\end{tabular}
}
\caption{ Dataset statistics (post-processing) for training. Neg\%: Proportion of input instances with no event occurrences. Events per row: Number of event triggers per input instance. Types per row: Number of unique event types per input instance. \#zs: Number of event types in test split not seen during training.
}
\label{tab:stats_proc}
\end{table}

\paragraph{RAMS} This dataset, created by \citet{ebner-etal-2020-multi}, is primarily geared towards the task of multi-sentence argument linking. 
The annotated argument roles are in a 5-sentence window around the related event trigger. 
In its native form, the dataset is geared towards multi-sentence argument role linking. 
Using the original configuration allows us to test the efficacy of our model on the multi-sentence level. 
Furthermore, on the sentence level, the dataset is imbalanced: 77\% of the sentences contain no events. 
Training a model on this incentivizes event occurrence detection over ED. 

\paragraph{MLEE}
This biomedical ED corpus by \citet{pyysalo_ohta_miwa_cho_tsujii_ananiadou_2012} is taken from PubMed abstracts centered around tissue-level and organ-level processes. 
The majority of the datasets used in this work are Event Extraction (EE) datasets, maintaining the scope of possible extensions of the proposed reformulation and multi-tasking approach to EE.

\section{Experiments and Results}



\subsection{Experimental Setup}
We use the generative T5 base (220M) model, a Transformer-based model. 


\paragraph{Hyperparameters} GPU: 2x NVIDIA GTX1080 GPUs. Maximum sequence length  1024 for multi-sentence input for 512 for sentence-level input. All models are trained for 50 epochs, with a batch size of 1. For beam search decoding, we use 50 beams.

To compare the efficacy of our method fairly with established baselines, we evaluate our predictions by converting them to token-level labels. 
We evaluate ED on two-level event type labels for RAMS and WikiEvents. 


\subsection{Results}

\paragraph{RAMS} As shown in Table \ref{tab:results_rams}, we achieve a 71.33\% F-1 score, which surpasses GPTEDOT by 6.2\%. 
 Furthermore, the difference between precision and recall is drastically lower than the competing non-generative, indicating that our model is less biased, and more robust.


\paragraph{WikiEvents}
As there are no existing event detection baselines on this dataset, we use single-task ED sequence generation performance as a baseline. 
This helps contextualize the benefits of our proposed prompted multi-task learning approach. 
We establish the benchmark performance of 60.7\% F-1 score on this dataset. 
The single-task and EDM3 micro F-1 and weighted F-1 scores can be found in Table \ref{tab:results_wiki}.
This is the model performance over the entire dataset, including negative instances, where false positives may occur.
On evaluating solely over the sentences with at least one event, we observe that the performance increases to 65.67\%. See App. \ref{EDM3_vs_singletask} for the example.

\begin{table}[t!]
\resizebox{\columnwidth}{!}{%
\begin{tabular}{l|ccc}
\toprule
\multicolumn{1}{l|}{\textbf{Model}} & \multicolumn{1}{c}{\textbf{P}} & \textbf{R} & \textbf{F-1} \\ \midrule
DMBERT \cite{wang-etal-2019-adversarial-training}                               & 62.6                                 & 44.0            & 51.7              \\
GatedGCN \cite{lai-etal-2020-event}                             & 66.5                                 & 59.0            & 62.5              \\ 
GPTEDOT \cite{Veyseh2021UnleashGP}                              & 55.5                                 & \textbf{78.6}            & 65.1              \\ \midrule
\textbf{EDM3}                   & \textbf{71.6}                                & 71.0           & \textbf{71.3}             \\ \bottomrule
\end{tabular}%
}
\caption{Results on RAMS. All previous models are sentence-level BERT-based models.}
\label{tab:results_rams}
\end{table}
\begin{table}[t!]
\centering
\fontsize{6pt}{8pt}\selectfont
\resizebox{\columnwidth}{!}
{
    \begin{tabular}{l|cccc}
\toprule
\textbf{Model}       & \multicolumn{1}{c}{\textbf{P}} & \multicolumn{1}{c}{\textbf{R}} & \multicolumn{1}{c}{\textbf{F-1}} & \multicolumn{1}{c}{\textbf{W-1}} \\ \midrule
Single-task & 60.0                  & 49.6                  & 54.3                    & 52.1                    \\
EDM3        & 60.8                  & 60.6                  & 60.7                    & \textbf{59.4}           \\ \bottomrule
\end{tabular}
}
\caption{Results on WikiEvents. W-1: Weighted F-1 \%}
\label{tab:results_wiki}
\end{table}
\begin{table}[t!]
\fontsize{8.5pt}{\baselineskip}\selectfont 
\color{black}
\resizebox{\columnwidth}{!}{%
\begin{tabular}{l|ccc|c}
\toprule
\textbf{Model}     & \multicolumn{1}{c}{\textbf{P}} & \multicolumn{1}{c}{\textbf{R}} & \multicolumn{1}{c}{\textbf{F-1}} & \multicolumn{1}{|c}{\textbf{F-1*}} \\ \midrule
SaliencyED \cite{liu-etal-2022-saliency}        & \textbf{64.9}                   & \textbf{69.4}                   & \textbf{67.1}                     & 60.3                               \\ \midrule
\textbf{EDM3} & 60.1                            & 65.5                            & 62.7                              & 58.1                      \\ \bottomrule
\end{tabular}%
}
\caption{Results on MAVEN. All results are on the publicly-available dev split. F-1*: Macro F-1 \%}
\label{tab:results_maven}
\end{table}
\begin{table}[!ht]
\centering
\resizebox{\columnwidth}{!}{%
\begin{tabular}{l|ccc}
\toprule
\textbf{Model}                                                & \textbf{P} & \textbf{R} & \textbf{F-1} \\ \midrule
SVM2 \cite{Zhou2015ASL} \textbf{*}                                                        & 72.2          & 82.3          & 76.9            \\ 
Two-stage \cite{7947109} \textbf{*}                                                   & 79.2          & 80.3          & 79.8            \\ \midrule
EANNP \cite{Nie2015EmbeddingAP}                                                         & 71.0          & \textbf{84.6}          & 77.2            \\ 

\begin{tabular}[c]{@{}l@{}}LSTM + CRF (w/o TL) \\ \end{tabular} & 81.6          & 74.3          & 77.8            \\ 
\begin{tabular}[c]{@{}l@{}}LSTM + CRF \cite{article}\\ \end{tabular}  & 81.8          & 77.7          & 79.7            \\ 
BiLSTM + Att \cite{2022A}                                                  & \textbf{82.0}          & 78.0          & \textbf{79.9}            \\ \midrule
EDM3                                                     & 75.9          & 80.4          & 78.1            \\ \bottomrule
\end{tabular}%
}
\caption{Results on MLEE dataset. \textbf{*} indicates models which require engineering hand-crafted features. All neural-network based models in this table use dependency-based embeddings specific to biomedical texts. w/o TL: results when 4 biomedical datasets are not used for transfer learning.}
\label{tab:results_mlee}
\end{table}
\begin{table}[t!]
\centering
\resizebox{\columnwidth}{!}{%
\begin{tabular}{@{}l|cc|cc|cc@{}}
\toprule
\multirow{2}{*}{\textbf{Dataset}} &
  \multicolumn{2}{c|}{\textbf{Single-task}} &
  \multicolumn{2}{c|}{\textbf{EDM3 (tags)}} &
  \multicolumn{2}{c}{\textbf{EDM3 (instr)}} \\ \cline{2-7} 
               & \multicolumn{1}{c}{All}   & Pos   & \multicolumn{1}{c}{All}   & Pos   & \multicolumn{1}{c}{All}            & Pos            \\ \hline
\textbf{MLEE}  & \multicolumn{1}{l}{71.07} & 72.20 & \multicolumn{1}{l}{74.57} & 75.82 & \multicolumn{1}{l}{\textbf{77.09}} & \textbf{78.45} \\ 
\textbf{RAMS}  & \multicolumn{1}{l}{63.21} & 63.21 & \multicolumn{1}{l}{67.66} & 67.66 & \multicolumn{1}{l}{\textbf{69.53}} & \textbf{69.53} \\ 
\textbf{MAVEN} & \multicolumn{1}{l}{58.10} & 59.18 & \multicolumn{1}{l}{62.29} & 63.56 & \multicolumn{1}{l}{\textbf{62.40}} & \textbf{63.66} \\ 
\textbf{WikiEvents} &
  \multicolumn{1}{l}{54.31} &
  58.47 &
  \multicolumn{1}{l}{56.77} &
  61.35 &
  \multicolumn{1}{l}{\textbf{58.71}} &
  \textbf{64.31} \\ \bottomrule
\end{tabular}%
}
\caption{Results on all datasets. Single-task: Event Detection results. EDM3 (tags): training with EI and EC tasks on the same dataset. EDM3 (instr): incorporating instructional prompts. All denotes performance on all input instances. Pos denotes performance on only event-containing instances.}
\label{tab:results_tagins}
\end{table}

\paragraph{MAVEN} 
We obtain a maximum F-1 score of 62.66\%, as seen in Table \ref{tab:results_maven}. 
While this score is below the existing best performance on this dataset, the class imbalance in the MAVEN dataset contributes to a lower micro F-1 score. 
This is shown by the fact that our model has a competitive macro F-1 score (58.1\% versus 60.3\%), indicating relatively better performance on sparsely populated classes. 
Further analysis of low-resource settings with examples can be found in the Analysis section of this work.
As shown in Table \ref{tab:results_mwt}, our model shows significant advantages in performing ED on complex instances, such as events with multi-class and multi-word event triggers which occur most frequently in this dataset as compared to others. 
To the best of our knowledge, we are the first to explicitly explore this facet of ED on MAVEN in the Analysis section. 




\paragraph{MLEE}
We distinguish between 2 sets of approaches for biomedical event detection as shown in Table \ref{tab:results_mlee}. 
The former set includes approaches that are comparatively labour-intensive, requiring the creation of handcrafted features for these tasks. 
The second set of models includes neural network-based models that use domain-specific embeddings obtained by parsing Pubmed or Medline abstracts. 
Even without domain-specific embeddings, our approach achieves 78.1\% F-1 score which is competitive with more sophisticated and domain-specific approaches.
We observe that our model also has higher recall (80.4\%) than the majority of the neural network-based approaches. More results are discussed in App. \ref{app:alternative_metrics}. 

\section{Analysis}
In this work, we conduct various experiments to assess the performance of our model over different scenarios. 



\subsection{Multi-tasking over EI and EC improves performance over ED}
Our hypothesis, that including EI and EC improves ED performance, is supported by results in Table \ref{tab:results_tagins}. 
EI helps identify multi-word triggers and event triggers that are missed in the single-task setting, while EC helps identify multi-class triggers. 
Even without instructional prompts, EDM3 improves performance by at least 3\% over single-tasking for all datasets. 
This can be attributed to the success of the subtask-level multi-tasking paradigm, with the improved performance due to the knowledge obtained by training the model over EI and EC in addition to the primary task of ED.
In the interests of a fair comparison, we use a greedy decoding scheme for all experiments conducted along this line of inquiry. 
Table \ref{tab:results_tagins} documents the metrics for single-task and multi-task models over all datasets. 
Examples demonstrating observably improved ED performance can be found in the appendix \S \ref{EDM3_vs_singletask}.



\subsection{Diversity is key to effective instructional prompts}
In this study, we investigate the impact of example diversity in instructional prompts on the performance of a task. 
Previous research suggests that prompts that consist of the task definition and two examples are optimal \cite{wang-etal-2022-super}. 
To examine the effect of example diversity, we include a biomedical example in the instructional prompt. 
Our findings indicate that examples, even from a different domain, can provide transferable knowledge.
The addition of a domain-relevant example results in the best performance on MLEE, achieving a score of 77.43\% before beam search decoding and 78.09\% after decoding. 
Moreover, the performance on general domain datasets improves with the inclusion of the biomedical example. 




\begin{table}[t]
\centering
\resizebox{\columnwidth}{!}{%
\begin{tabular}{l|cc|cc}
\toprule
\multirow{2}{*}{\textbf{Dataset}} & \multicolumn{2}{c|}{\textbf{Multi-word triggers}} & \multicolumn{2}{c}{\textbf{Multi-class triggers}} \\ \cmidrule{2-5} 
                    & \multicolumn{1}{c|}{\%instances}   & \%rows        & \multicolumn{1}{c|}{\%instances}   & \%rows        \\ \midrule
\textbf{RAMS}       & \multicolumn{1}{c|}{3.38} & 2.89 & \multicolumn{1}{c|}{\textbf{3.97}}          & \textbf{3.72}          \\ 
\textbf{MAVEN}      & \multicolumn{1}{c|}{\textbf{3.42}}          & \textbf{7.39}          & \multicolumn{1}{c|}{0.06} & 0.13 \\ 
\textbf{WikiEvents} & \multicolumn{1}{c|}{2.86}          & 2.18          & \multicolumn{1}{c|}{0}          & 0          \\ \bottomrule
\end{tabular}%
}
\caption{Statistics on multi-word and multi-class triggers in all datasets. \%instances: the \% of total triggers present. \%rows: the \% of all input instances that contain at least 1 multi-word or multi-class trigger.}
\label{tab:multi_trigs}
\end{table}
\begin{table}[t]
\centering
\fontsize{6pt}{8pt}\selectfont
\resizebox{\columnwidth}{!}{%
\begin{tabular}{lccc}
\toprule
\multirow{2}{*}{\textbf{Dataset}} & \multicolumn{2}{c}{\textbf{\#mwt}} & \multirow{2}{*}{\textbf{EM acc \%}} \\ \cmidrule{2-3}
           & \multicolumn{1}{c|}{Train} & Test &       \\ \midrule
MAVEN      & \multicolumn{1}{c|}{2442}  & 633  & 90.84 \\ 
RAMS       & \multicolumn{1}{c|}{228}   & 20   & 88.89 \\ 
WikiEvents & \multicolumn{1}{c|}{127}   & 18   & 44.44 \\ \bottomrule
\end{tabular}%
}
\caption{Results on multi-word triggers. \#mwt: number of multi-word triggers in training and testing data. EM acc \%: exact match accuracy, i.e. percentage of multi-word triggers in test data predicted by our model.}
\label{tab:results_mwt}
\end{table}

\subsection{Negative instances hamper ED performance}
From the dataset statistics in Table \ref{tab:stats_proc}, we see that the WikiEvents dataset has close to 54\% instances that have no annotated events, i.e. negative instances. 
We hypothesize that this detracts from the model's ability to discern relevant events and their types, and instead emphasizes the binary classification task of identifying event presence.
We analyze the effect of negative examples further experimentally (Table \ref{tab:results_tagins}). 
The consistent trend of higher Pos scores indicates that, given a sentence, our approach is better at extracting its events accurately as opposed to identifying whether it contains an event. 

The difference between both metrics is stark in the case of WikiEvents. 
We observe increased performance (60.71\% to 65.67\% after beam search decoding) over WikiEvents, which is significantly higher than what we observe on other datasets.
From further analysis, we find that training on only positive examples improves the ED performance on event sentences by nearly 5\%. 
Furthermore, despite the fact that MAVEN has 168 event types and WikiEvents has only 49 (Table \ref{tab:stats_overview}), the ED performance on MAVEN (62.4\%) is higher than on WikiEvents (58.7\%). 
This indicates that rather than the complexity of the ED task, the distribution of positive and negative instances may hamper the model's ability to perform the task.

%
%

\begin{figure}[t!]
	\centering
	\includegraphics[width= \linewidth, height= 2.3cm]{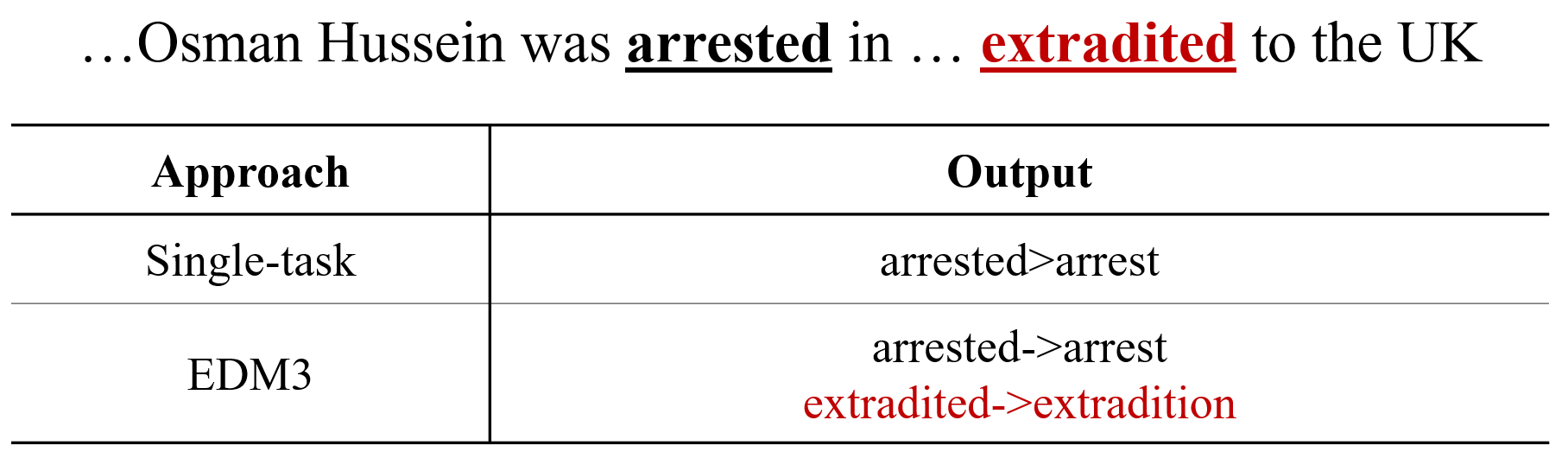}
	\caption{EDM3 capturing the event type \textit{extradition}, which has only 11 annotated instances in MAVEN.}
	\label{fig:example_low_resource}
\end{figure}

\subsection{EDM3 is well-suited to low-resource scenarios}
The majority of instances in MAVEN deal with a subset of its 168 event types. \citet{https://doi.org/10.48550/arxiv.2204.12456} show that 18\% of all event types have less than 100 annotated instances, making them hard to learn and identify. 
For example, the event types, \textit{Breathing} and \textit{Extradition}, have less than 20 annotated train instances in more than 8K training sentences (6 and 11 annotated triggers, respectively). 
Despite this, we see the model accurately identifies all triggers in test data that are of these event types (see Figure \ref{fig:example_low_resource}), achieving 100\% testing precision on both, and 100\% and 80\% micro F-1 score respectively.


\subsection{Successful identification of multi-word triggers}

Token classification is inadequate for accurately measuring the performance of ED on real-world datasets with multi-word event triggers, which comprise a significant portion of triggers (3.42\% in MAVEN and 3.38\% in RAMS) as shown in Table \ref{tab:multi_trigs}.
Treating multi-word triggers as individual tokens can yield misleading results, as many triggers only represent the event type when the entire phrase is annotated. For example, for the trigger phrase "took place", labeling only either "took" or "place" would be incorrect: the individual words are semantically distinct from the meaning of the whole phrase, and individually denote different event types. 

To evaluate our model's performance on multi-word trigger phrases, we calculate exact match accuracy for all multi-word triggers. 
We achieve nearly 91\% and 89\% on MAVEN and RAMS, respectively (Table \ref{tab:results_mwt}).
Although the WikiEvents dataset has fewer multi-word triggers, our model achieves a respectable performance, with partially predicted triggers ("assault", "in touch") often being semantically similar to the gold annotations ("the assault", "been in touch").
For trigger phrases where partial predictions are semantically unequal to complete predictions, such as "took place" and "set off," our model still performs well.







\begin{figure}[t!]
	\centering
	\includegraphics[width= \linewidth, height= 3.3 cm]{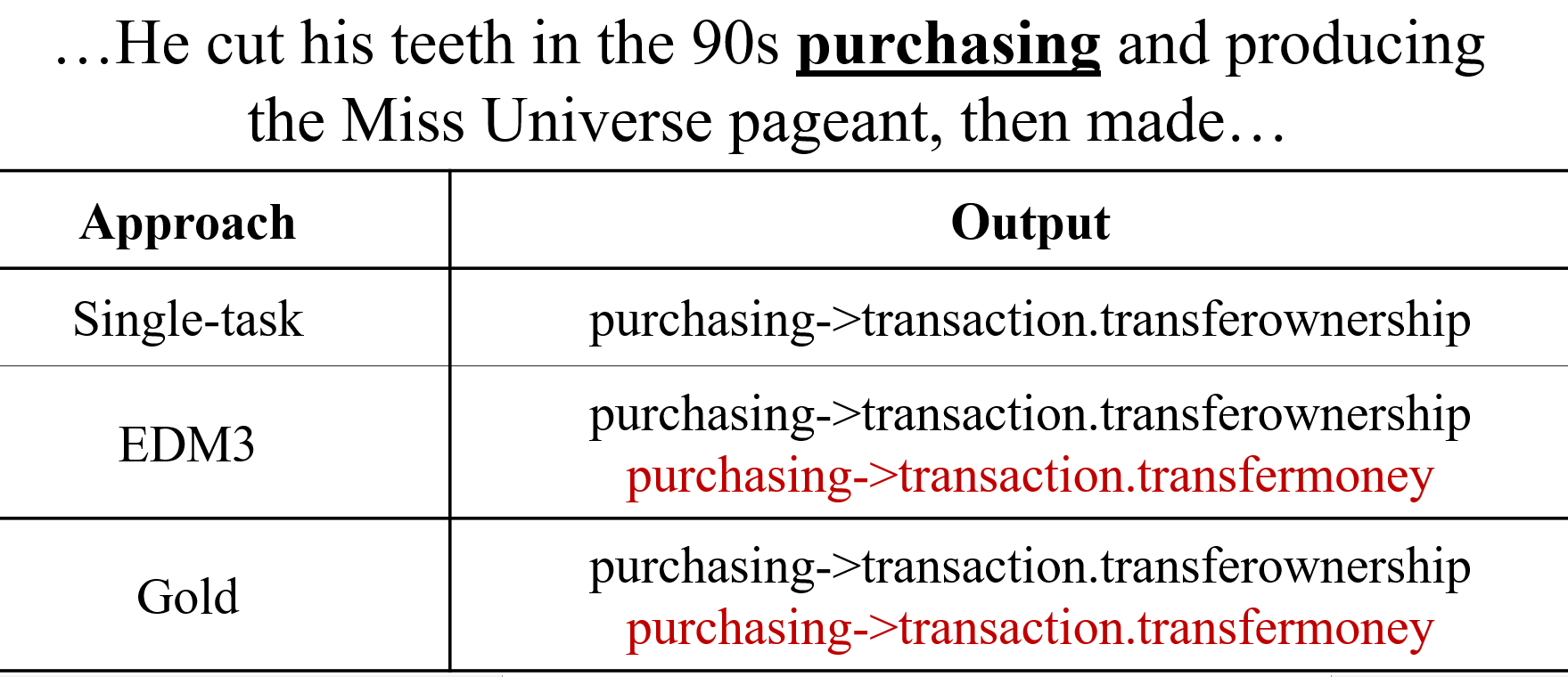}
	\caption{EDM3 improving prediction on multi-class triggers. In the single-task setting, only one sense of the event trigger is identified. EDM3 accurately extracts all senses of the given multi-class trigger.}
    \label{fig:mcl_1}
\end{figure}

\subsection{Successful classification of multi-class triggers}
In a real-world ED scenario, event triggers may function as triggers of multiple event types within the same context. 
We observe these most commonly in RAMS, where nearly 4\% of all event triggers are classified as triggers of multiple event types (see Table \ref{tab:multi_trigs}).
For example, \textbf{purchasing} in Figure \ref{fig:mcl_1} triggers two distinct types of transaction events. 
\textit{transferownership} is an event type with arguments such as previous and current owner, while \textit{transfermoney} requires the \textit{amount} as an argument.
To accurately detect these events, it is necessary to capture all the senses of a particular trigger. 

Existing token classification methods are not well-suited to this task as they perform and evaluate event detection as multi-class classification rather than multi-label classification. 
Our approach of multi-tasking over subtasks, specifically, training the model over EC, enables the model to predict multi-class triggers. 
We use prediction accuracy to evaluate the model's performance on multi-class triggers. The accuracy is evaluated as 50\% over a particular multi-class trigger if we predict one of two event types that it triggers in that input instance. 
We find that the average prediction accuracy is close to 61\% on the RAMS dataset, indicating that the model can capture most of the senses in which each multi-class trigger functions.


\subsection{Multi-sentence context is crucial to ED}

Consider the examples from the WikiEvents dataset.

\noindent\textit{Example 1}: The whole building has \textbf{collapsed}.

\noindent\textit{Example 2}: He chose \textbf{destruction}.



\noindent In Example 1, our model extracts the token in bold as a relevant event trigger and classifies it as an event of the type $artifactexistence$ with the subtype $damage-destroy-disable-dismantle$. 
However, upon closer examination, we find that this example is taken from a document that primarily focuses on events of the type $conflict.attack$, with \textbf{bombing} and \textbf{explosion} being the annotated event triggers. 
Therefore, \textbf{collapsed} can be seen as an auxiliary event, and the model should predict the sentence as NONE. 
Conversely, in Example 2, our model classifies the sentence as NONE, indicating no salient event was found. 
However, the following sentences in the same document provide the necessary context to demonstrate that destruction is, in fact, the salient event in this case. 
The gold annotation identifies \textbf{destruction} as a trigger of event type \textit{artifactexistence} with the subtype \textit{damage-destroy-disable-dismantle}.

This shows us that sentences tagged NONE may nevertheless have salient events predicted by the model, but are tagged NONE because, in the original multi-sentence context, the salient event in the sentence is less important than events that are the subject of the passage. 
It is difficult for our model to judge the saliency of an event without the semantic context of its document, and the relevance of other events in its vicinity. 
This is why it is vital to include multi-sentence or document-level context, as sentence-level information can be misleading in the broader context.

\section{Conclusion}

In this paper, we propose a domain-agnostic generative approach to the Event Detection task that demonstrates the effectiveness of breaking down complex generation tasks into subtasks. 
Our method leverages a multi-tasking strategy that incorporates instructional prompts to improve model performance on imbalanced data and complex event instances. 
Our analysis shows an improvement in F-1 score over single-task performance, supporting our main hypothesis viz. the effectiveness of breaking down complex generation tasks into subtasks that can support model learning on the primary task. 
Furthermore, our results highlight the potential for generative models in traditionally discriminative tasks like ED, paving the way for future advancements in the field.

\section*{Limitations}
Our work demonstrates a prompted and generative approach on a single task, Event Detection, which can be easily adapted to other information retrieval tasks. 
However, the model faces relative difficulty in distinguishing non-event sentences, which could be addressed by implementing a binary classification system. 
In addition, including contextual information could help identify trigger candidates better. 
Our decoding scheme can also be improved for better recall without negatively impacting precision. 
Furthermore, there is a possibility of improving prompt quality further by analyzing the number and scope of examples required to achieve the best prompted performance. 
Finally, integrating domain knowledge could improve event-type classification, and we encourage future researchers to explore this area. 
Despite these limitations, our work provides a strong foundation for generative, instructional prompt-based frameworks for end-to-end Event Extraction and opens up exciting avenues for future research in this area.




\section*{Acknowledgement}
We thank the Research Computing (RC) at Arizona State University (ASU) for providing computing resources for experiments.

\bibliography{anthology,custom}

\begin{thebibliography}{104}
\expandafter\ifx\csname natexlab\endcsname\relax\def\natexlab#1{#1}\fi

\bibitem[{Ahn(2006)}]{ahn-2006-stages}
David Ahn. 2006.
\newblock \href {https://aclanthology.org/W06-0901} {The stages of event
  extraction}.
\newblock In \emph{Proceedings of the Workshop on Annotating and Reasoning
  about Time and Events}, pages 1--8, Sydney, Australia. Association for
  Computational Linguistics.

\bibitem[{Allamanis et~al.(2017)Allamanis, Brockschmidt, and
  Khademi}]{allamanis2017learning}
Miltiadis Allamanis, Marc Brockschmidt, and Mahmoud Khademi. 2017.
\newblock Learning to represent programs with graphs.
\newblock \emph{arXiv preprint arXiv:1711.00740}.

\bibitem[{Alon et~al.(2018)Alon, Zilberstein, Levy, and
  Yahav}]{Alon2018code2vecLD}
Uri Alon, Meital Zilberstein, Omer Levy, and Eran Yahav. 2018.
\newblock code2vec: learning distributed representations of code.
\newblock \emph{Proceedings of the ACM on Programming Languages}, 3:1 -- 29.

\bibitem[{Balog et~al.(2016)Balog, Gaunt, Brockschmidt, Nowozin, and
  Tarlow}]{balog2016deepcoder}
Matej Balog, Alexander~L Gaunt, Marc Brockschmidt, Sebastian Nowozin, and
  Daniel Tarlow. 2016.
\newblock Deepcoder: Learning to write programs.
\newblock \emph{arXiv preprint arXiv:1611.01989}.

\bibitem[{Boros et~al.(2021)Boros, Moreno, and
  Doucet}]{DBLP:journals/corr/abs-2104-06969}
Emanuela Boros, Jos{\'{e}}~G. Moreno, and Antoine Doucet. 2021.
\newblock \href {http://arxiv.org/abs/2104.06969} {Event detection as question
  answering with entity information}.
\newblock \emph{CoRR}, abs/2104.06969.

\bibitem[{Bronstein et~al.(2015)Bronstein, Dagan, Li, Ji, and
  Frank}]{bronstein-etal-2015-seed}
Ofer Bronstein, Ido Dagan, Qi~Li, Heng Ji, and Anette Frank. 2015.
\newblock \href {https://doi.org/10.3115/v1/P15-2061} {Seed-based event trigger
  labeling: How far can event descriptions get us?}
\newblock In \emph{Proceedings of the 53rd Annual Meeting of the Association
  for Computational Linguistics and the 7th International Joint Conference on
  Natural Language Processing (Volume 2: Short Papers)}, pages 372--376,
  Beijing, China. Association for Computational Linguistics.

\bibitem[{Caruana(1997)}]{10.1023/A:1007379606734}
Rich Caruana. 1997.
\newblock \href {https://doi.org/10.1023/A:1007379606734} {Multitask learning}.
\newblock \emph{Mach. Learn.}, 28(1):41–75.

\bibitem[{Chen and Ng(2012)}]{inproceedings}
Chen Chen and Vincent Ng. 2012.
\newblock Joint modeling for chinese event extraction with rich linguistic
  features.
\newblock pages 529--544.

\bibitem[{Chen(2019)}]{article}
Yifei Chen. 2019.
\newblock \href {https://doi.org/10.1186/s12859-019-3030-z} {Multiple-level
  biomedical event trigger recognition with transfer learning}.
\newblock \emph{BMC Bioinformatics}, 20.

\bibitem[{Chen et~al.(2017)Chen, Liu, Zhang, Liu, and
  Zhao}]{chen-etal-2017-automatically}
Yubo Chen, Shulin Liu, Xiang Zhang, Kang Liu, and Jun Zhao. 2017.
\newblock \href {https://doi.org/10.18653/v1/P17-1038} {Automatically labeled
  data generation for large scale event extraction}.
\newblock In \emph{Proceedings of the 55th Annual Meeting of the Association
  for Computational Linguistics (Volume 1: Long Papers)}, pages 409--419,
  Vancouver, Canada. Association for Computational Linguistics.

\bibitem[{Chen et~al.(2015)Chen, Xu, Liu, Zeng, and
  Zhao}]{chen-etal-2015-event}
Yubo Chen, Liheng Xu, Kang Liu, Daojian Zeng, and Jun Zhao. 2015.
\newblock \href {https://doi.org/10.3115/v1/P15-1017} {Event extraction via
  dynamic multi-pooling convolutional neural networks}.
\newblock In \emph{Proceedings of the 53rd Annual Meeting of the Association
  for Computational Linguistics and the 7th International Joint Conference on
  Natural Language Processing (Volume 1: Long Papers)}, pages 167--176,
  Beijing, China. Association for Computational Linguistics.

\bibitem[{Cheng and Erk(2018)}]{cheng-erk-2018-implicit}
Pengxiang Cheng and Katrin Erk. 2018.
\newblock \href {https://doi.org/10.18653/v1/N18-1076} {Implicit argument
  prediction with event knowledge}.
\newblock In \emph{Proceedings of the 2018 Conference of the North {A}merican
  Chapter of the Association for Computational Linguistics: Human Language
  Technologies, Volume 1 (Long Papers)}, pages 831--840, New Orleans,
  Louisiana. Association for Computational Linguistics.

\bibitem[{Crawshaw(2020)}]{https://doi.org/10.48550/arxiv.2009.09796}
Michael Crawshaw. 2020.
\newblock \href {https://doi.org/10.48550/ARXIV.2009.09796} {Multi-task
  learning with deep neural networks: A survey}.

\bibitem[{Das(2015)}]{Das2015ContextualCC}
Subhasis Das. 2015.
\newblock Contextual code completion using machine learning.

\bibitem[{Deng et~al.(2021)Deng, Zhang, Li, Hui, Huaixiao, Chen, Huang, and
  Chen}]{deng-etal-2021-ontoed}
Shumin Deng, Ningyu Zhang, Luoqiu Li, Chen Hui, Tou Huaixiao, Mosha Chen, Fei
  Huang, and Huajun Chen. 2021.
\newblock \href {https://doi.org/10.18653/v1/2021.acl-long.220} {{O}nto{ED}:
  Low-resource event detection with ontology embedding}.
\newblock In \emph{Proceedings of the 59th Annual Meeting of the Association
  for Computational Linguistics and the 11th International Joint Conference on
  Natural Language Processing (Volume 1: Long Papers)}, pages 2828--2839,
  Online. Association for Computational Linguistics.

\bibitem[{Du and Cardie(2020)}]{du-cardie-2020-event}
Xinya Du and Claire Cardie. 2020.
\newblock \href {https://doi.org/10.18653/v1/2020.emnlp-main.49} {Event
  extraction by answering (almost) natural questions}.
\newblock In \emph{Proceedings of the 2020 Conference on Empirical Methods in
  Natural Language Processing (EMNLP)}, pages 671--683, Online. Association for
  Computational Linguistics.

\bibitem[{Ebner et~al.(2020)Ebner, Xia, Culkin, Rawlins, and
  Van~Durme}]{ebner-etal-2020-multi}
Seth Ebner, Patrick Xia, Ryan Culkin, Kyle Rawlins, and Benjamin Van~Durme.
  2020.
\newblock \href {https://doi.org/10.18653/v1/2020.acl-main.718} {Multi-sentence
  argument linking}.
\newblock In \emph{Proceedings of the 58th Annual Meeting of the Association
  for Computational Linguistics}, pages 8057--8077, Online. Association for
  Computational Linguistics.

\bibitem[{Efrat and Levy(2020)}]{https://doi.org/10.48550/arxiv.2010.11982}
Avia Efrat and Omer Levy. 2020.
\newblock \href {https://doi.org/10.48550/ARXIV.2010.11982} {The turking test:
  Can language models understand instructions?}

\bibitem[{Feng et~al.(2020)Feng, Guo, Tang, Duan, Feng, Gong, Shou, Qin, Liu,
  Jiang, and Zhou}]{feng-etal-2020-codebert}
Zhangyin Feng, Daya Guo, Duyu Tang, Nan Duan, Xiaocheng Feng, Ming Gong, Linjun
  Shou, Bing Qin, Ting Liu, Daxin Jiang, and Ming Zhou. 2020.
\newblock \href {https://doi.org/10.18653/v1/2020.findings-emnlp.139}
  {{C}ode{BERT}: A pre-trained model for programming and natural languages}.
\newblock In \emph{Findings of the Association for Computational Linguistics:
  EMNLP 2020}, pages 1536--1547, Online. Association for Computational
  Linguistics.

\bibitem[{Ghaeini et~al.(2016)Ghaeini, Fern, Huang, and
  Tadepalli}]{ghaeini-etal-2016-event}
Reza Ghaeini, Xiaoli Fern, Liang Huang, and Prasad Tadepalli. 2016.
\newblock \href {https://doi.org/10.18653/v1/P16-2060} {Event nugget detection
  with forward-backward recurrent neural networks}.
\newblock In \emph{Proceedings of the 54th Annual Meeting of the Association
  for Computational Linguistics (Volume 2: Short Papers)}, pages 369--373,
  Berlin, Germany. Association for Computational Linguistics.

\bibitem[{Gupta et~al.(2021{\natexlab{a}})Gupta, Gulanikar, Kumar, and
  Neti}]{10.1007/978-3-030-87013-3_4}
Himanshu Gupta, Abhiram~Anand Gulanikar, Lov Kumar, and Lalita Bhanu~Murthy
  Neti. 2021{\natexlab{a}}.
\newblock Empirical analysis on effectiveness of nlp methods for predicting
  code smell.
\newblock In \emph{Computational Science and Its Applications -- ICCSA 2021},
  pages 43--53, Cham. Springer International Publishing.

\bibitem[{Gupta et~al.(2020)Gupta, Kulkarni, Kumar, and
  Murthy}]{10.1007/978-3-030-45778-5_27}
Himanshu Gupta, Tanmay~Girish Kulkarni, Lov Kumar, and Neti Lalita~Bhanu
  Murthy. 2020.
\newblock A novel approach towards analysis of attacker behavior in ddos
  attacks.
\newblock In \emph{Machine Learning for Networking}, pages 392--402, Cham.
  Springer International Publishing.

\bibitem[{Gupta et~al.(2021{\natexlab{b}})Gupta, Kulkarni, Kumar, Neti, and
  Krishna}]{10.1007/978-3-030-75075-6_10}
Himanshu Gupta, Tanmay~Girish Kulkarni, Lov Kumar, Lalita Bhanu~Murthy Neti,
  and Aneesh Krishna. 2021{\natexlab{b}}.
\newblock An empirical study on predictability of software code smell using
  deep learning models.
\newblock In \emph{Advanced Information Networking and Applications}, pages
  120--132, Cham. Springer International Publishing.

\bibitem[{Gupta et~al.(2019)Gupta, Kumar, and Neti}]{8877082}
Himanshu Gupta, Lov Kumar, and Lalita Bhanu~Murthy Neti. 2019.
\newblock \href {https://doi.org/10.1109/IEMECONX.2019.8877082} {An empirical
  framework for code smell prediction using extreme learning machine}.
\newblock In \emph{2019 9th Annual Information Technology, Electromechanical
  Engineering and Microelectronics Conference (IEMECON)}, pages 189--195.

\bibitem[{Gupta et~al.(2021{\natexlab{c}})Gupta, Misra, Kumar, and
  Murthy}]{10.1007/978-3-030-69143-1_18}
Himanshu Gupta, Sanjay Misra, Lov Kumar, and N.~L.~Bhanu Murthy.
  2021{\natexlab{c}}.
\newblock An empirical study to investigate data sampling techniques for
  improving code-smell prediction using imbalanced data.
\newblock In \emph{Information and Communication Technology and Applications},
  pages 220--233, Cham. Springer International Publishing.

\bibitem[{Gupta et~al.(2021{\natexlab{d}})Gupta, Verma, Kumar, Mishra, Agrawal,
  Badugu, and Bhatt}]{gupta2021context}
Himanshu Gupta, Shreyas Verma, Tarun Kumar, Swaroop Mishra, Tamanna Agrawal,
  Amogh Badugu, and Himanshu~Sharad Bhatt. 2021{\natexlab{d}}.
\newblock Context-ner: Contextual phrase generation at scale.
\newblock \emph{arXiv preprint arXiv:2109.08079}.

\bibitem[{Gupta and Ji(2009)}]{gupta-ji-2009-predicting}
Prashant Gupta and Heng Ji. 2009.
\newblock \href {https://aclanthology.org/P09-2093} {Predicting unknown time
  arguments based on cross-event propagation}.
\newblock In \emph{Proceedings of the {ACL}-{IJCNLP} 2009 Conference Short
  Papers}, pages 369--372, Suntec, Singapore. Association for Computational
  Linguistics.

\bibitem[{Hase and Bansal(2021)}]{hase2021models}
Peter Hase and Mohit Bansal. 2021.
\newblock \href {http://arxiv.org/abs/2102.02201} {When can models learn from
  explanations? a formal framework for understanding the roles of explanation
  data}.

\bibitem[{He et~al.(2018{\natexlab{a}})He, Li, Liu, Yu, and Meng}]{7947109}
Xinyu He, Lishuang Li, Yang Liu, Xiaoming Yu, and Jun Meng. 2018{\natexlab{a}}.
\newblock \href {https://doi.org/10.1109/TCBB.2017.2715016} {A two-stage
  biomedical event trigger detection method integrating feature selection and
  word embeddings}.
\newblock \emph{IEEE/ACM Transactions on Computational Biology and
  Bioinformatics}, 15(4):1325--1332.

\bibitem[{He et~al.(2018{\natexlab{b}})He, Li, Wan, Song, Meng, and
  Wang}]{8621217}
Xinyu He, Lishuang Li, Jia Wan, Dingxin Song, Jun Meng, and Zhanjie Wang.
  2018{\natexlab{b}}.
\newblock \href {https://doi.org/10.1109/BIBM.2018.8621217} {Biomedical event
  trigger detection based on bilstm integrating attention mechanism and
  sentence vector}.
\newblock In \emph{2018 IEEE International Conference on Bioinformatics and
  Biomedicine (BIBM)}, pages 651--654.

\bibitem[{He et~al.(2022)He, Tai, Lu, Huang, and Ren}]{2022A}
Xinyu He, Ping Tai, Hongbin Lu, Xin Huang, and Yonggong Ren. 2022.
\newblock A biomedical event extraction method based on fine-grained and
  attention mechanism.
\newblock \emph{BMC Bioinformatics}, 23(1):1--17.

\bibitem[{Hong et~al.(2011)Hong, Zhang, Ma, Yao, Zhou, and
  Zhu}]{hong-etal-2011-using}
Yu~Hong, Jianfeng Zhang, Bin Ma, Jianmin Yao, Guodong Zhou, and Qiaoming Zhu.
  2011.
\newblock \href {https://aclanthology.org/P11-1113} {Using cross-entity
  inference to improve event extraction}.
\newblock In \emph{Proceedings of the 49th Annual Meeting of the Association
  for Computational Linguistics: Human Language Technologies}, pages
  1127--1136, Portland, Oregon, USA. Association for Computational Linguistics.

\bibitem[{Huang and Riloff(2021)}]{Huang_Riloff_2021}
Ruihong Huang and Ellen Riloff. 2021.
\newblock \href {https://doi.org/10.1609/aaai.v26i1.8354} {Modeling textual
  cohesion for event extraction}.
\newblock \emph{Proceedings of the AAAI Conference on Artificial Intelligence},
  26(1):1664--1670.

\bibitem[{Husain et~al.(2019)Husain, Wu, Gazit, Allamanis, and
  Brockschmidt}]{Husain2019CodeSearchNetCE}
Hamel Husain, Hongqi Wu, Tiferet Gazit, Miltiadis Allamanis, and Marc
  Brockschmidt. 2019.
\newblock Codesearchnet challenge: Evaluating the state of semantic code
  search.
\newblock \emph{ArXiv}, abs/1909.09436.

\bibitem[{Ji and Grishman(2008)}]{ji-grishman-2008-refining}
Heng Ji and Ralph Grishman. 2008.
\newblock \href {https://aclanthology.org/P08-1030} {Refining event extraction
  through cross-document inference}.
\newblock In \emph{Proceedings of ACL-08: HLT}, pages 254--262, Columbus, Ohio.
  Association for Computational Linguistics.

\bibitem[{Kanhabua and Anand(2016)}]{10.1145/2911451.2914805}
Nattiya Kanhabua and Avishek Anand. 2016.
\newblock \href {https://doi.org/10.1145/2911451.2914805} {Temporal information
  retrieval}.
\newblock In \emph{Proceedings of the 39th International ACM SIGIR Conference
  on Research and Development in Information Retrieval}, SIGIR '16, page
  1235–1238, New York, NY, USA. Association for Computing Machinery.

\bibitem[{Kehinde et~al.(2022)Kehinde, Adeniyi, Ogundokun, Gupta, and
  Misra}]{10.1007/978-981-16-8892-8_46}
Adeniyi~Jide Kehinde, Abidemi~Emmanuel Adeniyi, Roseline~Oluwaseun Ogundokun,
  Himanshu Gupta, and Sanjay Misra. 2022.
\newblock Prediction of students' performance with artificial neural network
  using demographic traits.
\newblock In \emph{Recent Innovations in Computing}, pages 613--624, Singapore.
  Springer Singapore.

\bibitem[{Lai and Nguyen(2019)}]{DBLP:journals/corr/abs-1910-11368}
Viet~Dac Lai and Thien~Huu Nguyen. 2019.
\newblock \href {http://arxiv.org/abs/1910.11368} {Extending event detection to
  new types with learning from keywords}.
\newblock \emph{CoRR}, abs/1910.11368.

\bibitem[{Lai et~al.(2020{\natexlab{a}})Lai, Nguyen, and
  Nguyen}]{lai-etal-2020-event}
Viet~Dac Lai, Tuan~Ngo Nguyen, and Thien~Huu Nguyen. 2020{\natexlab{a}}.
\newblock \href {https://doi.org/10.18653/v1/2020.emnlp-main.435} {Event
  detection: Gate diversity and syntactic importance scores for graph
  convolution neural networks}.
\newblock In \emph{Proceedings of the 2020 Conference on Empirical Methods in
  Natural Language Processing (EMNLP)}, pages 5405--5411, Online. Association
  for Computational Linguistics.

\bibitem[{Lai et~al.(2020{\natexlab{b}})Lai, Nguyen, and Nguyen}]{lai2020event}
Viet~Dac Lai, Tuan~Ngo Nguyen, and Thien~Huu Nguyen. 2020{\natexlab{b}}.
\newblock Event detection: Gate diversity and syntactic importance scoresfor
  graph convolution neural networks.
\newblock \emph{arXiv preprint arXiv:2010.14123}.

\bibitem[{Levy and Goldberg(2014)}]{levy-goldberg-2014-dependency}
Omer Levy and Yoav Goldberg. 2014.
\newblock \href {https://doi.org/10.3115/v1/P14-2050} {Dependency-based word
  embeddings}.
\newblock In \emph{Proceedings of the 52nd Annual Meeting of the Association
  for Computational Linguistics (Volume 2: Short Papers)}, pages 302--308,
  Baltimore, Maryland. Association for Computational Linguistics.

\bibitem[{Li et~al.(2017)Li, Wang, Lyu, and King}]{li2017code}
Jian Li, Yue Wang, Michael~R Lyu, and Irwin King. 2017.
\newblock Code completion with neural attention and pointer networks.
\newblock \emph{arXiv preprint arXiv:1711.09573}.

\bibitem[{Li et~al.(2013{\natexlab{a}})Li, Zhu, and
  Zhou}]{li-etal-2013-argument}
Peifeng Li, Qiaoming Zhu, and Guodong Zhou. 2013{\natexlab{a}}.
\newblock \href {https://aclanthology.org/P13-1145} {Argument inference from
  relevant event mentions in {C}hinese argument extraction}.
\newblock In \emph{Proceedings of the 51st Annual Meeting of the Association
  for Computational Linguistics (Volume 1: Long Papers)}, pages 1477--1487,
  Sofia, Bulgaria. Association for Computational Linguistics.

\bibitem[{Li et~al.(2013{\natexlab{b}})Li, Ji, and Huang}]{li-etal-2013-joint}
Qi~Li, Heng Ji, and Liang Huang. 2013{\natexlab{b}}.
\newblock \href {https://aclanthology.org/P13-1008} {Joint event extraction via
  structured prediction with global features}.
\newblock In \emph{Proceedings of the 51st Annual Meeting of the Association
  for Computational Linguistics (Volume 1: Long Papers)}, pages 73--82, Sofia,
  Bulgaria. Association for Computational Linguistics.

\bibitem[{Li et~al.(2021)Li, Ji, and Han}]{Li2021DocumentLevelEA}
Sha Li, Heng Ji, and Jiawei Han. 2021.
\newblock Document-level event argument extraction by conditional generation.
\newblock \emph{ArXiv}, abs/2104.05919.

\bibitem[{Liao and Grishman(2010)}]{liao-grishman-2010-using}
Shasha Liao and Ralph Grishman. 2010.
\newblock \href {https://aclanthology.org/P10-1081} {Using document level
  cross-event inference to improve event extraction}.
\newblock In \emph{Proceedings of the 48th Annual Meeting of the Association
  for Computational Linguistics}, pages 789--797, Uppsala, Sweden. Association
  for Computational Linguistics.

\bibitem[{Liao and Grishman(2011)}]{liao-grishman-2011-acquiring}
Shasha Liao and Ralph Grishman. 2011.
\newblock \href {https://aclanthology.org/R11-1002} {Acquiring topic features
  to improve event extraction: in pre-selected and balanced collections}.
\newblock In \emph{Proceedings of the International Conference Recent Advances
  in Natural Language Processing 2011}, pages 9--16, Hissar, Bulgaria.
  Association for Computational Linguistics.

\bibitem[{Lin et~al.(2020)Lin, Ji, Huang, and Wu}]{lin-etal-2020-joint}
Ying Lin, Heng Ji, Fei Huang, and Lingfei Wu. 2020.
\newblock \href {https://doi.org/10.18653/v1/2020.acl-main.713} {A joint neural
  model for information extraction with global features}.
\newblock In \emph{Proceedings of the 58th Annual Meeting of the Association
  for Computational Linguistics}, pages 7999--8009, Online. Association for
  Computational Linguistics.

\bibitem[{Liu et~al.(2020)Liu, Chen, Liu, Bi, and Liu}]{liu-etal-2020-event}
Jian Liu, Yubo Chen, Kang Liu, Wei Bi, and Xiaojiang Liu. 2020.
\newblock \href {https://doi.org/10.18653/v1/2020.emnlp-main.128} {Event
  extraction as machine reading comprehension}.
\newblock In \emph{Proceedings of the 2020 Conference on Empirical Methods in
  Natural Language Processing (EMNLP)}, pages 1641--1651, Online. Association
  for Computational Linguistics.

\bibitem[{Liu et~al.(2022)Liu, Chen, and Xu}]{liu-etal-2022-saliency}
Jian Liu, Yufeng Chen, and Jinan Xu. 2022.
\newblock \href {https://doi.org/10.18653/v1/2022.acl-long.313} {Saliency as
  evidence: Event detection with trigger saliency attribution}.
\newblock In \emph{Proceedings of the 60th Annual Meeting of the Association
  for Computational Linguistics (Volume 1: Long Papers)}, pages 4573--4585,
  Dublin, Ireland. Association for Computational Linguistics.

\bibitem[{Liu et~al.(2021)Liu, Yuan, Fu, Jiang, Hayashi, and
  Neubig}]{DBLP:journals/corr/abs-2107-13586}
Pengfei Liu, Weizhe Yuan, Jinlan Fu, Zhengbao Jiang, Hiroaki Hayashi, and
  Graham Neubig. 2021.
\newblock \href {http://arxiv.org/abs/2107.13586} {Pre-train, prompt, and
  predict: {A} systematic survey of prompting methods in natural language
  processing}.
\newblock \emph{CoRR}, abs/2107.13586.

\bibitem[{Liu et~al.(2017)Liu, Chen, Liu, and Zhao}]{liu-etal-2017-exploiting}
Shulin Liu, Yubo Chen, Kang Liu, and Jun Zhao. 2017.
\newblock \href {https://doi.org/10.18653/v1/P17-1164} {Exploiting argument
  information to improve event detection via supervised attention mechanisms}.
\newblock In \emph{Proceedings of the 55th Annual Meeting of the Association
  for Computational Linguistics (Volume 1: Long Papers)}, pages 1789--1798,
  Vancouver, Canada. Association for Computational Linguistics.

\bibitem[{Liu et~al.(2016)Liu, Liu, He, and Zhao}]{liu2016probabilistic}
Shulin Liu, Kang Liu, Shizhu He, and Jun Zhao. 2016.
\newblock A probabilistic soft logic based approach to exploiting latent and
  global information in event classification.
\newblock In \emph{Proceedings of the AAAI Conference on Artificial
  Intelligence}, volume~30.

\bibitem[{Liu et~al.(2018)Liu, Luo, and Huang}]{liu-etal-2018-jointly}
Xiao Liu, Zhunchen Luo, and Heyan Huang. 2018.
\newblock \href {https://doi.org/10.18653/v1/D18-1156} {Jointly multiple events
  extraction via attention-based graph information aggregation}.
\newblock In \emph{Proceedings of the 2018 Conference on Empirical Methods in
  Natural Language Processing}, pages 1247--1256, Brussels, Belgium.
  Association for Computational Linguistics.

\bibitem[{Lourie et~al.(2021)Lourie, Bras, Bhagavatula, and
  Choi}]{lourie2021unicorn}
Nicholas Lourie, Ronan~Le Bras, Chandra Bhagavatula, and Yejin Choi. 2021.
\newblock \href {http://arxiv.org/abs/2103.13009} {Unicorn on rainbow: A
  universal commonsense reasoning model on a new multitask benchmark}.

\bibitem[{Lu et~al.(2019)Lu, Lin, Han, and Sun}]{lu-etal-2019-distilling}
Yaojie Lu, Hongyu Lin, Xianpei Han, and Le~Sun. 2019.
\newblock \href {https://doi.org/10.18653/v1/P19-1429} {Distilling
  discrimination and generalization knowledge for event detection via
  delta-representation learning}.
\newblock In \emph{Proceedings of the 57th Annual Meeting of the Association
  for Computational Linguistics}, pages 4366--4376, Florence, Italy.
  Association for Computational Linguistics.

\bibitem[{Lu et~al.(2021)Lu, Lin, Xu, Han, Tang, Li, Sun, Liao, and
  Chen}]{lu-etal-2021-text2event}
Yaojie Lu, Hongyu Lin, Jin Xu, Xianpei Han, Jialong Tang, Annan Li, Le~Sun,
  Meng Liao, and Shaoyi Chen. 2021.
\newblock \href {https://doi.org/10.18653/v1/2021.acl-long.217}
  {{T}ext2{E}vent: Controllable sequence-to-structure generation for end-to-end
  event extraction}.
\newblock In \emph{Proceedings of the 59th Annual Meeting of the Association
  for Computational Linguistics and the 11th International Joint Conference on
  Natural Language Processing (Volume 1: Long Papers)}, pages 2795--2806,
  Online. Association for Computational Linguistics.

\bibitem[{Lyu et~al.(2021)Lyu, Zhang, Sulem, and Roth}]{lyu2021zero}
Qing Lyu, Hongming Zhang, Elior Sulem, and Dan Roth. 2021.
\newblock Zero-shot event extraction via transfer learning: Challenges and
  insights.
\newblock In \emph{Proceedings of the 59th Annual Meeting of the Association
  for Computational Linguistics and the 11th International Joint Conference on
  Natural Language Processing (Volume 2: Short Papers)}, pages 322--332.

\bibitem[{McClosky et~al.(2011)McClosky, Surdeanu, and
  Manning}]{mcclosky-etal-2011-event}
David McClosky, Mihai Surdeanu, and Christopher Manning. 2011.
\newblock \href {https://aclanthology.org/P11-1163} {Event extraction as
  dependency parsing}.
\newblock In \emph{Proceedings of the 49th Annual Meeting of the Association
  for Computational Linguistics: Human Language Technologies}, pages
  1626--1635, Portland, Oregon, USA. Association for Computational Linguistics.

\bibitem[{Mishra et~al.(2022)Mishra, Khashabi, Baral, Choi, and
  Hajishirzi}]{mishra2022reframing}
Swaroop Mishra, Daniel Khashabi, Chitta Baral, Yejin Choi, and Hannaneh
  Hajishirzi. 2022.
\newblock \href {http://arxiv.org/abs/2109.07830} {Reframing instructional
  prompts to gptk's language}.

\bibitem[{Nakayama(2018)}]{seqeval}
Hiroki Nakayama. 2018.
\newblock \href {https://github.com/chakki-works/seqeval} {{seqeval}: A python
  framework for sequence labeling evaluation}.
\newblock Software available from https://github.com/chakki-works/seqeval.

\bibitem[{Nguyen et~al.(2016)Nguyen, Cho, and
  Grishman}]{nguyen-etal-2016-joint-event}
Thien~Huu Nguyen, Kyunghyun Cho, and Ralph Grishman. 2016.
\newblock \href {https://doi.org/10.18653/v1/N16-1034} {Joint event extraction
  via recurrent neural networks}.
\newblock In \emph{Proceedings of the 2016 Conference of the North {A}merican
  Chapter of the Association for Computational Linguistics: Human Language
  Technologies}, pages 300--309, San Diego, California. Association for
  Computational Linguistics.

\bibitem[{Nguyen and Grishman(2015)}]{nguyen-grishman-2015-event}
Thien~Huu Nguyen and Ralph Grishman. 2015.
\newblock \href {https://doi.org/10.3115/v1/P15-2060} {Event detection and
  domain adaptation with convolutional neural networks}.
\newblock In \emph{Proceedings of the 53rd Annual Meeting of the Association
  for Computational Linguistics and the 7th International Joint Conference on
  Natural Language Processing (Volume 2: Short Papers)}, pages 365--371,
  Beijing, China. Association for Computational Linguistics.

\bibitem[{Nie et~al.(2015)Nie, Rong, Zhang, Ouyang, and
  Xiong}]{Nie2015EmbeddingAP}
Yifan Nie, Wenge Rong, Yiyuan Zhang, Yuanxin Ouyang, and Zhang Xiong. 2015.
\newblock Embedding assisted prediction architecture for event trigger
  identification.
\newblock \emph{Journal of bioinformatics and computational biology}, 13
  3:1541001.

\bibitem[{Ogundokun et~al.(2022)Ogundokun, Misra, Sadiku, Gupta, Damasevicius,
  and Maskeliunas}]{10.1007/978-981-16-8892-8_29}
Roseline~Oluwaseun Ogundokun, Sanjay Misra, Peter~Ogirima Sadiku, Himanshu
  Gupta, Robertas Damasevicius, and Rytis Maskeliunas. 2022.
\newblock Computational intelligence approaches for heart disease detection.
\newblock In \emph{Recent Innovations in Computing}, pages 385--395, Singapore.
  Springer Singapore.

\bibitem[{Paolini et~al.(2021)Paolini, Athiwaratkun, Krone, Ma, Achille,
  Anubhai, dos Santos, Xiang, and Soatto}]{tanl}
Giovanni Paolini, Ben Athiwaratkun, Jason Krone, Jie Ma, Alessandro Achille,
  Rishita Anubhai, Cicero~Nogueira dos Santos, Bing Xiang, and Stefano Soatto.
  2021.
\newblock Structured prediction as translation between augmented natural
  languages.
\newblock In \emph{9th International Conference on Learning Representations,
  {ICLR} 2021}.

\bibitem[{Parmar et~al.(2022)Parmar, Mishra, Purohit, Luo, Murad, and
  Baral}]{parmar2022inboxbart}
Mihir Parmar, Swaroop Mishra, Mirali Purohit, Man Luo, M.~Hassan Murad, and
  Chitta Baral. 2022.
\newblock \href {http://arxiv.org/abs/2204.07600} {In-boxbart: Get instructions
  into biomedical multi-task learning}.

\bibitem[{Patwardhan and Riloff(2009)}]{patwardhan-riloff-2009-unified}
Siddharth Patwardhan and Ellen Riloff. 2009.
\newblock \href {https://aclanthology.org/D09-1016} {A unified model of phrasal
  and sentential evidence for information extraction}.
\newblock In \emph{Proceedings of the 2009 Conference on Empirical Methods in
  Natural Language Processing}, pages 151--160, Singapore. Association for
  Computational Linguistics.

\bibitem[{Pyysalo et~al.(2013)Pyysalo, Ginter, Moen, Salakoski, and
  Ananiadou}]{8b865ebe417a41fbacb7241efe6a5490}
S~Pyysalo, F~Ginter, H~Moen, T~Salakoski, and S~Ananiadou. 2013.
\newblock Distributional semantics resources for biomedical text processing.
\newblock In \emph{Proceedings of LBM 2013}, pages 39--44.

\bibitem[{Pyysalo et~al.(2012)Pyysalo, Ohta, Miwa, Cho, Tsujii, and
  Ananiadou}]{pyysalo_ohta_miwa_cho_tsujii_ananiadou_2012}
Sampo Pyysalo, Tomoko Ohta, Makoto Miwa, Han-Cheol Cho, Jun'ichi Tsujii, and
  Sophia Ananiadou. 2012.
\newblock \href {https://doi.org/10.1093/bioinformatics/bts407} {Event
  extraction across multiple levels of biological organization}.
\newblock \emph{Bioinformatics}, 28(18):i575–i581.

\bibitem[{Raffel et~al.(2020)Raffel, Shazeer, Roberts, Lee, Narang, Matena,
  Zhou, Li, and Liu}]{raffel2020exploring}
Colin Raffel, Noam Shazeer, Adam Roberts, Katherine Lee, Sharan Narang, Michael
  Matena, Yanqi Zhou, Wei Li, and Peter~J. Liu. 2020.
\newblock \href {http://arxiv.org/abs/1910.10683} {Exploring the limits of
  transfer learning with a unified text-to-text transformer}.

\bibitem[{Riedel and McCallum(2011{\natexlab{a}})}]{riedel-mccallum-2011-fast}
Sebastian Riedel and Andrew McCallum. 2011{\natexlab{a}}.
\newblock \href {https://aclanthology.org/D11-1001} {Fast and robust joint
  models for biomedical event extraction}.
\newblock In \emph{Proceedings of the 2011 Conference on Empirical Methods in
  Natural Language Processing}, pages 1--12, Edinburgh, Scotland, UK.
  Association for Computational Linguistics.

\bibitem[{Riedel and
  McCallum(2011{\natexlab{b}})}]{riedel-mccallum-2011-robust}
Sebastian Riedel and Andrew McCallum. 2011{\natexlab{b}}.
\newblock \href {https://aclanthology.org/W11-1807} {Robust biomedical event
  extraction with dual decomposition and minimal domain adaptation}.
\newblock In \emph{Proceedings of {B}io{NLP} Shared Task 2011 Workshop}, pages
  46--50, Portland, Oregon, USA. Association for Computational Linguistics.

\bibitem[{Sanh et~al.(2021)Sanh, Webson, Raffel, Bach, Sutawika, Alyafeai,
  Chaffin, Stiegler, Scao, Raja et~al.}]{sanh2021multitask}
Victor Sanh, Albert Webson, Colin Raffel, Stephen~H Bach, Lintang Sutawika,
  Zaid Alyafeai, Antoine Chaffin, Arnaud Stiegler, Teven~Le Scao, Arun Raja,
  et~al. 2021.
\newblock Multitask prompted training enables zero-shot task generalization.
\newblock \emph{arXiv preprint arXiv:2110.08207}.

\bibitem[{Scaria et~al.(2023)Scaria, Gupta, Sawant, Mishra, and
  Baral}]{scaria2023instructabsa}
Kevin Scaria, Himanshu Gupta, Saurabh~Arjun Sawant, Swaroop Mishra, and Chitta
  Baral. 2023.
\newblock Instructabsa: Instruction learning for aspect based sentiment
  analysis.
\newblock \emph{arXiv preprint arXiv:2302.08624}.

\bibitem[{Sha et~al.(2018)Sha, Qian, Chang, and Sui}]{Sha_Qian_Chang_Sui_2018}
Lei Sha, Feng Qian, Baobao Chang, and Zhifang Sui. 2018.
\newblock \href {https://doi.org/10.1609/aaai.v32i1.12034} {Jointly extracting
  event triggers and arguments by dependency-bridge rnn and tensor-based
  argument interaction}.
\newblock \emph{Proceedings of the AAAI Conference on Artificial Intelligence},
  32(1).

\bibitem[{Si et~al.(2022)Si, Peng, Li, Xu, and Li}]{si2022generating}
Jinghui Si, Xutan Peng, Chen Li, Haotian Xu, and Jianxin Li. 2022.
\newblock \href {http://arxiv.org/abs/2110.04525} {Generating disentangled
  arguments with prompts: A simple event extraction framework that works}.

\bibitem[{Souza~Costa et~al.(2020)Souza~Costa, Gottschalk, and
  Demidova}]{souza2020event}
Tarc{\'\i}sio Souza~Costa, Simon Gottschalk, and Elena Demidova. 2020.
\newblock Event-qa: A dataset for event-centric question answering over
  knowledge graphs.
\newblock In \emph{Proceedings of the 29th ACM international conference on
  information \& knowledge management}, pages 3157--3164.

\bibitem[{Tong et~al.(2020)Tong, Xu, Wang, Cao, Hou, Li, and
  Xie}]{tong-etal-2020-improving}
Meihan Tong, Bin Xu, Shuai Wang, Yixin Cao, Lei Hou, Juanzi Li, and Jun Xie.
  2020.
\newblock \href {https://doi.org/10.18653/v1/2020.acl-main.522} {Improving
  event detection via open-domain trigger knowledge}.
\newblock In \emph{Proceedings of the 58th Annual Meeting of the Association
  for Computational Linguistics}, pages 5887--5897, Online. Association for
  Computational Linguistics.

\bibitem[{V~S S~Patchigolla et~al.(2017)V~S S~Patchigolla, Sahu, and
  Anand}]{v-s-s-patchigolla-etal-2017-biomedical}
Rahul V~S S~Patchigolla, Sunil Sahu, and Ashish Anand. 2017.
\newblock \href {https://doi.org/10.18653/v1/W17-2340} {Biomedical event
  trigger identification using bidirectional recurrent neural network based
  models}.
\newblock In \emph{{B}io{NLP} 2017}, pages 316--321, Vancouver, Canada,.
  Association for Computational Linguistics.

\bibitem[{Vaswani et~al.(2017)Vaswani, Shazeer, Parmar, Uszkoreit, Jones,
  Gomez, Kaiser, and Polosukhin}]{NIPS2017_3f5ee243}
Ashish Vaswani, Noam Shazeer, Niki Parmar, Jakob Uszkoreit, Llion Jones,
  Aidan~N Gomez, \L~ukasz Kaiser, and Illia Polosukhin. 2017.
\newblock \href
  {https://proceedings.neurips.cc/paper/2017/file/3f5ee243547dee91fbd053c1c4a845aa-Paper.pdf}
  {Attention is all you need}.
\newblock In \emph{Advances in Neural Information Processing Systems},
  volume~30. Curran Associates, Inc.

\bibitem[{Venugopal et~al.(2014)Venugopal, Chen, Gogate, and
  Ng}]{venugopal-etal-2014-relieving}
Deepak Venugopal, Chen Chen, Vibhav Gogate, and Vincent Ng. 2014.
\newblock \href {https://doi.org/10.3115/v1/D14-1090} {Relieving the
  computational bottleneck: Joint inference for event extraction with
  high-dimensional features}.
\newblock In \emph{Proceedings of the 2014 Conference on Empirical Methods in
  Natural Language Processing ({EMNLP})}, pages 831--843, Doha, Qatar.
  Association for Computational Linguistics.

\bibitem[{Veyseh et~al.(2021)Veyseh, Lai, Dernoncourt, and
  Nguyen}]{Veyseh2021UnleashGP}
Amir Pouran~Ben Veyseh, Viet~Dac Lai, Franck Dernoncourt, and Thien~Huu Nguyen.
  2021.
\newblock Unleash gpt-2 power for event detection.
\newblock In \emph{ACL}.

\bibitem[{Vijayakumar et~al.(2018)Vijayakumar, Mohta, Polozov, Batra, Jain, and
  Gulwani}]{Vijayakumar2018NeuralGuidedDS}
Ashwin~J. Vijayakumar, Abhishek Mohta, Oleksandr Polozov, Dhruv Batra, Prateek
  Jain, and Sumit Gulwani. 2018.
\newblock Neural-guided deductive search for real-time program synthesis from
  examples.
\newblock \emph{ArXiv}, abs/1804.01186.

\bibitem[{Wadden et~al.(2019)Wadden, Wennberg, Luan, and
  Hajishirzi}]{wadden-etal-2019-entity}
David Wadden, Ulme Wennberg, Yi~Luan, and Hannaneh Hajishirzi. 2019.
\newblock \href {https://doi.org/10.18653/v1/D19-1585} {Entity, relation, and
  event extraction with contextualized span representations}.
\newblock In \emph{Proceedings of the 2019 Conference on Empirical Methods in
  Natural Language Processing and the 9th International Joint Conference on
  Natural Language Processing (EMNLP-IJCNLP)}, pages 5784--5789, Hong Kong,
  China. Association for Computational Linguistics.

\bibitem[{Wang et~al.(2017)Wang, Wang, Lin, Zhang, Yang, and Xu}]{Wang2017AMD}
Anran Wang, Jian Wang, Hongfei Lin, Jianhai Zhang, Zhihao Yang, and Kan Xu.
  2017.
\newblock A multiple distributed representation method based on neural network
  for biomedical event extraction.
\newblock \emph{BMC Medical Informatics and Decision Making}, 17.

\bibitem[{Wang and Cohen(2009)}]{wang-cohen-2009-character}
Richard~C. Wang and William~W. Cohen. 2009.
\newblock \href {https://aclanthology.org/D09-1156} {Character-level analysis
  of semi-structured documents for set expansion}.
\newblock In \emph{Proceedings of the 2009 Conference on Empirical Methods in
  Natural Language Processing}, pages 1503--1512, Singapore. Association for
  Computational Linguistics.

\bibitem[{Wang et~al.(2021)Wang, Yu, Chang, Sun, and
  Huang}]{DBLP:journals/corr/abs-2110-07476}
Sijia Wang, Mo~Yu, Shiyu Chang, Lichao Sun, and Lifu Huang. 2021.
\newblock \href {http://arxiv.org/abs/2110.07476} {Query and extract: Refining
  event extraction as type-oriented binary decoding}.
\newblock \emph{CoRR}, abs/2110.07476.

\bibitem[{Wang et~al.(2022{\natexlab{a}})Wang, Yu, and Huang}]{wang2022art}
Sijia Wang, Mo~Yu, and Lifu Huang. 2022{\natexlab{a}}.
\newblock \href {http://arxiv.org/abs/2204.07241} {The art of prompting: Event
  detection based on type specific prompts}.

\bibitem[{Wang et~al.(2019)Wang, Han, Liu, Sun, and
  Li}]{wang-etal-2019-adversarial-training}
Xiaozhi Wang, Xu~Han, Zhiyuan Liu, Maosong Sun, and Peng Li. 2019.
\newblock \href {https://doi.org/10.18653/v1/N19-1105} {Adversarial training
  for weakly supervised event detection}.
\newblock In \emph{Proceedings of the 2019 Conference of the North {A}merican
  Chapter of the Association for Computational Linguistics: Human Language
  Technologies, Volume 1 (Long and Short Papers)}, pages 998--1008,
  Minneapolis, Minnesota. Association for Computational Linguistics.

\bibitem[{Wang et~al.(2020)Wang, Wang, Han, Jiang, Han, Liu, Li, Li, Lin, and
  Zhou}]{wang-etal-2020-maven}
Xiaozhi Wang, Ziqi Wang, Xu~Han, Wangyi Jiang, Rong Han, Zhiyuan Liu, Juanzi
  Li, Peng Li, Yankai Lin, and Jie Zhou. 2020.
\newblock \href {https://doi.org/10.18653/v1/2020.emnlp-main.129} {{MAVEN}: {A}
  {M}assive {G}eneral {D}omain {E}vent {D}etection {D}ataset}.
\newblock In \emph{Proceedings of the 2020 Conference on Empirical Methods in
  Natural Language Processing (EMNLP)}, pages 1652--1671, Online. Association
  for Computational Linguistics.

\bibitem[{Wang et~al.(2022{\natexlab{b}})Wang, Mishra, Alipoormolabashi, Kordi,
  Mirzaei, Naik, Ashok, Dhanasekaran, Arunkumar, Stap, Pathak, Karamanolakis,
  Lai, Purohit, Mondal, Anderson, Kuznia, Doshi, Pal, Patel, Moradshahi,
  Parmar, Purohit, Varshney, Kaza, Verma, Puri, Karia, Doshi, Sampat, Mishra,
  Reddy~A, Patro, Dixit, and Shen}]{wang-etal-2022-super}
Yizhong Wang, Swaroop Mishra, Pegah Alipoormolabashi, Yeganeh Kordi, Amirreza
  Mirzaei, Atharva Naik, Arjun Ashok, Arut~Selvan Dhanasekaran, Anjana
  Arunkumar, David Stap, Eshaan Pathak, Giannis Karamanolakis, Haizhi Lai,
  Ishan Purohit, Ishani Mondal, Jacob Anderson, Kirby Kuznia, Krima Doshi,
  Kuntal~Kumar Pal, Maitreya Patel, Mehrad Moradshahi, Mihir Parmar, Mirali
  Purohit, Neeraj Varshney, Phani~Rohitha Kaza, Pulkit Verma, Ravsehaj~Singh
  Puri, Rushang Karia, Savan Doshi, Shailaja~Keyur Sampat, Siddhartha Mishra,
  Sujan Reddy~A, Sumanta Patro, Tanay Dixit, and Xudong Shen.
  2022{\natexlab{b}}.
\newblock \href {https://aclanthology.org/2022.emnlp-main.340}
  {Super-{N}atural{I}nstructions: Generalization via declarative instructions
  on 1600+ {NLP} tasks}.
\newblock In \emph{Proceedings of the 2022 Conference on Empirical Methods in
  Natural Language Processing}, pages 5085--5109, Abu Dhabi, United Arab
  Emirates. Association for Computational Linguistics.

\bibitem[{Webson and Pavlick(2022)}]{webson2022promptbased}
Albert Webson and Ellie Pavlick. 2022.
\newblock \href {http://arxiv.org/abs/2109.01247} {Do prompt-based models
  really understand the meaning of their prompts?}

\bibitem[{Wei et~al.(2022)Wei, Bosma, Zhao, Guu, Yu, Lester, Du, Dai, and
  Le}]{wei2022finetuned}
Jason Wei, Maarten Bosma, Vincent~Y. Zhao, Kelvin Guu, Adams~Wei Yu, Brian
  Lester, Nan Du, Andrew~M. Dai, and Quoc~V. Le. 2022.
\newblock \href {http://arxiv.org/abs/2109.01652} {Finetuned language models
  are zero-shot learners}.

\bibitem[{Xie et~al.(2022)Xie, Wu, Shi, Zhong, Scholak, Yasunaga, Wu, Zhong,
  Yin, Wang, Zhong, Wang, Li, Boyle, Ni, Yao, Radev, Xiong, Kong, Zhang, Smith,
  Zettlemoyer, and Yu}]{xie2022unifiedskg}
Tianbao Xie, Chen~Henry Wu, Peng Shi, Ruiqi Zhong, Torsten Scholak, Michihiro
  Yasunaga, Chien-Sheng Wu, Ming Zhong, Pengcheng Yin, Sida~I. Wang, Victor
  Zhong, Bailin Wang, Chengzu Li, Connor Boyle, Ansong Ni, Ziyu Yao, Dragomir
  Radev, Caiming Xiong, Lingpeng Kong, Rui Zhang, Noah~A. Smith, Luke
  Zettlemoyer, and Tao Yu. 2022.
\newblock \href {http://arxiv.org/abs/2201.05966} {Unifiedskg: Unifying and
  multi-tasking structured knowledge grounding with text-to-text language
  models}.

\bibitem[{Xu et~al.(2018)Xu, Qian, Mei, Huang, and Liu}]{10.1145/3287075}
Mengwei Xu, Feng Qian, Qiaozhu Mei, Kang Huang, and Xuanzhe Liu. 2018.
\newblock \href {https://doi.org/10.1145/3287075} {Deeptype: On-device deep
  learning for input personalization service with minimal privacy concern}.
\newblock \emph{Proc. ACM Interact. Mob. Wearable Ubiquitous Technol.}, 2(4).

\bibitem[{Yang et~al.(2019)Yang, Feng, Qiao, Kan, and
  Li}]{yang-etal-2019-exploring}
Sen Yang, Dawei Feng, Linbo Qiao, Zhigang Kan, and Dongsheng Li. 2019.
\newblock \href {https://doi.org/10.18653/v1/P19-1522} {Exploring pre-trained
  language models for event extraction and generation}.
\newblock In \emph{Proceedings of the 57th Annual Meeting of the Association
  for Computational Linguistics}, pages 5284--5294, Florence, Italy.
  Association for Computational Linguistics.

\bibitem[{Ye and Ren(2021)}]{ye2021learning}
Qinyuan Ye and Xiang Ren. 2021.
\newblock \href {http://arxiv.org/abs/2101.00420} {Learning to generate
  task-specific adapters from task description}.

\bibitem[{Yin et~al.(2018)Yin, Neubig, Allamanis, Brockschmidt, and
  Gaunt}]{Yin2018LearningTR}
Pengcheng Yin, Graham Neubig, Miltiadis Allamanis, Marc Brockschmidt, and
  Alexander~L. Gaunt. 2018.
\newblock Learning to represent edits.
\newblock \emph{ArXiv}, abs/1810.13337.

\bibitem[{Zhang et~al.(2021)Zhang, Wang, and Roth}]{zhang-etal-2021-zero}
Hongming Zhang, Haoyu Wang, and Dan Roth. 2021.
\newblock \href {https://doi.org/10.18653/v1/2021.findings-acl.114}
  {{Z}ero-shot {L}abel-aware {E}vent {T}rigger and {A}rgument
  {C}lassification}.
\newblock In \emph{Findings of the Association for Computational Linguistics:
  ACL-IJCNLP 2021}, pages 1331--1340, Online. Association for Computational
  Linguistics.

\bibitem[{Zhang et~al.(2022)Zhang, Ingale, Shabir, Li, Shi, and
  Wang}]{https://doi.org/10.48550/arxiv.2204.12456}
Wenlong Zhang, Bhagyashree Ingale, Hamza Shabir, Tianyi Li, Tian Shi, and Ping
  Wang. 2022.
\newblock \href {https://doi.org/10.48550/ARXIV.2204.12456} {Event detection
  explorer: An interactive tool for event detection exploration}.

\bibitem[{Zhao et~al.(2021)Zhao, Khashabi, Khot, Sabharwal, and
  Chang}]{zhao2021ethicaladvice}
Jieyu Zhao, Daniel Khashabi, Tushar Khot, Ashish Sabharwal, and Kai-Wei Chang.
  2021.
\newblock \href {http://arxiv.org/abs/2106.01465} {Ethical-advice taker: Do
  language models understand natural language interventions?}

\bibitem[{Zhong et~al.(2021)Zhong, Lee, Zhang, and Klein}]{zhong2021adapting}
Ruiqi Zhong, Kristy Lee, Zheng Zhang, and Dan Klein. 2021.
\newblock \href {http://arxiv.org/abs/2104.04670} {Adapting language models for
  zero-shot learning by meta-tuning on dataset and prompt collections}.

\bibitem[{Zhou and Zhong(2015)}]{Zhou2015ASL}
Deyu Zhou and Dayou Zhong. 2015.
\newblock A semi-supervised learning framework for biomedical event extraction
  based on hidden topics.
\newblock \emph{Artificial intelligence in medicine}, 64 1:51--8.

\end{thebibliography}
\bibliographystyle{acl_natbib}

\clearpage

\section*{Appendix}
\appendix

\section{Extended Related Work}
\label{sec:Other_Related_Work}

LMs and Deep learning methods have been used for a plethora of downstream tasks for a long time \cite{Yin2018LearningTR,li2017code,Das2015ContextualCC,10.1007/978-3-030-45778-5_27, 10.1007/978-3-030-75075-6_10,10.1007/978-3-030-69143-1_18,10.1007/978-3-030-87013-3_4,Husain2019CodeSearchNetCE,feng-etal-2020-codebert,Vijayakumar2018NeuralGuidedDS}.
Several recent works have leveraged NLP methods and simple sampling methods for different downstream results \cite{10.1145/3287075,Alon2018code2vecLD,allamanis2017learning,balog2016deepcoder,10.1007/978-981-16-8892-8_29,10.1007/978-981-16-8892-8_46,8877082}.

\subsection{Event Detection and Identification}
\citet{ahn-2006-stages} was the first work to distinguish between the subtasks of Event Detection and Event Identification. Early ED \cite{gupta-ji-2009-predicting,liao-grishman-2010-using,ji-grishman-2008-refining,hong-etal-2011-using,liu2016probabilistic} frameworks used highly engineered features. \citet{inproceedings} proved that joint training allowed knowledge sharing and reduced error propagation along the ED pipeline. This denoted a shift from pipelined models \citep{ji-grishman-2008-refining,gupta-ji-2009-predicting,patwardhan-riloff-2009-unified,liao-grishman-2011-acquiring,mcclosky-etal-2011-event,Huang_Riloff_2021,li-etal-2013-argument} 
to joint training architectures \cite{riedel-mccallum-2011-robust,riedel-mccallum-2011-fast,li-etal-2013-joint,venugopal-etal-2014-relieving,chen-etal-2017-automatically,liu-etal-2017-exploiting}.

Neural network-based methods leveraged pretrained embeddings such as Word2Vec as features. \citet{nguyen-grishman-2015-event,chen-etal-2015-event} formulated ED as a token-classification problem and proved domain-adaptability. NN-based models \cite{chen-etal-2015-event} used a range of other architectures including RNNs \cite{ghaeini-etal-2016-event,Sha_Qian_Chang_Sui_2018}, LSTMs \cite{nguyen-etal-2016-joint-event}, and GCNs \citep{liu-etal-2018-jointly}.


\subsection{Prompt Engineering}
Using prompts and natural language instructions to augment input data and improve model learning is an active research area. The turking test \citep{https://doi.org/10.48550/arxiv.2010.11982} was proposed as a method to evaluate how well machine learning models can learn from instructions, akin to humans, on a range of tasks. Later works have investigated how well PLMs gain a semantic understanding of prompts 
\citep{webson2022promptbased,zhao2021ethicaladvice}. The instruction learning paradigm has been investigated in detail \citep{hase2021models,ye2021learning,mishra2022reframing}, especially in settings such as low-resource or zero-shot settings \citep{zhong2021adapting,sanh2021multitask,wei2022finetuned}. 
Several studies are present that show adding knowledge with instruction helps LMs understand the context better \cite{scaria2023instructabsa,gupta2021context} .



\subsection{Prompting for Generative ED}
Adding natural language prompts have shown promising results in improving performance in PLMs \citep{DBLP:journals/corr/abs-2107-13586}. PLMs are trained on general domain data, making them less suited to successful right-out-of-the-box applications on tasks which require domain-specific knowledge and contexts.  Prompt engineering is an active area of research across domains. 
Unlike previous prompt-based approaches \citep{wang2022art}, we do not create prompts solely within the scope of the ED task, such as event type-specific prompts. 
Adding natural language prompts have shown promising results in improving performance in PLMs \citep{DBLP:journals/corr/abs-2107-13586}. 
Unlike previous prompt-based approaches \citep{wang2022art}, we do not create prompts solely within the scope of the ED task, such as event type-specific prompts.  

\subsection{Previous State of the Art}

 \paragraph{RAMS} Existing baselines perform ED on sentence-level. We compare our multi-sentence ED performance with DMBERT \cite{wang-etal-2019-adversarial-training}, GatedGCN \cite{lai-etal-2020-event}, and GPTEDOT \cite{Veyseh2021UnleashGP}. All these models are BERT-based. The state-of-the-art model, GPTEDOT, leverages the multi-task learning paradigm similarly, however, owing to the limits of classification-based representations, only uses EI, and requires generation of data to augment existing examples. 

\paragraph{MAVEN}  The state-of-the-art model, \citep{wang2022art}, leverages the prompting paradigm to perform word classification for ED. This model requires significant prompt engineering for all 168 event types in the schema. However, as its performance is evaluated on the unavailable test split, with access to possible trigger candidates, we do not report its performance as a comparable baseline. SaliencyED \cite{liu-etal-2022-saliency} explicitly states that it performs trigger extraction for trigger words, with no mention made of multi-class, or the far more frequent, multi-word triggers.

\paragraph{MLEE} We distinguish between 2 sets of models for biomedical event detection. The former set of models include 2 SVM-based models \cite{pyysalo_ohta_miwa_cho_tsujii_ananiadou_2012,Zhou2015ASL}, and a pipelined two-stage model \cite{7947109}, which treats event identification and classification separately. 
These approaches are comparatively labour-intensive; they require the creation of handcrafted features for these tasks. The second set of models are neural network-based models. 
These include a CNN with embeddings encoding event type, POS labels and topic representation \cite{Wang2017AMD}, RNN with word and entity embeddings \cite{v-s-s-patchigolla-etal-2017-biomedical}, and LSTM-based models that integrate other biomedical datasets in order to perform transfer learning \cite{article}.
All existing baselines use pretrained embeddings and other language resources specifically for biomedical texts such as the resources published by \cite{8b865ebe417a41fbacb7241efe6a5490}. 
For example, LSTM \cite{8621217} and the state-of-the-art BiLSTM \cite{2022A}, like the majority of existing models, employs Word2vecf \cite{levy-goldberg-2014-dependency} to train dependency-based word embeddings. 
These embeddings are trained on Pubmed abstracts that are parsed using the Gdep parser: a dependency parse tool built for use on biomedical texts. 
Likewise, EANNP \cite{Nie2015EmbeddingAP} undertakes a similar approach to pretraining embeddings, but uses Medline abstracts instead.

\section{Alternative evaluation metric for sequence generation-based ED}
\label{app:alternative_metrics}

The majority of existing works treat this as a multi-class word classification problem, and all baselines, including generative methods, evaluate model results consistent with word classification metrics popularly used for NER tasks \cite{seqeval}. However, we see many multi-label trigger words. This becomes especially apparent in document level data, and in real world data, the same trigger may function as a trigger for multiple event types, with a different set of arguments corresponding to its role as each event type it triggers. This makes existing numbers misleading. 

As an alternative evaluation scheme, we treat sequence generation based ED as a sentence or multi-sentence level multilabel classification, where multi-word triggers are considered distinct labels. For this problem, we treat NONE as a possible label for a given input text. 

We calculate the metrics of precision, recall, and F-1 score using conventional formulae:
\newline\newline
\centerline {Precision (P) = $\frac{TP}{TP+FP}$}
\newline \newline
\centerline {Recall (R) = $\frac{TP}{TP+FN}$}
\newline \newline
\centerline{F-1 score = $2 \times \frac{P \times R}{P + R}$}
\newline 

where a prediction is counted as true positive only if both trigger span and predicted event type (including subtype) match gold annotations.
 
In addition to more accurate performance metrics over multi-class triggers, this provides a stricter metric to evaluate multi-word triggers, where partial predictions do not contribute to model performance. Using this metric also allows us to evaluate the discriminative performance of an ED model, i.e. the accuracy with which it can identify whether an input text contains an event or not. We implement this evaluation metric based on publicly-available code from another sequence generation model for ED. The results on entire test data as well as event and non-event sentences obtained using this metric are reported in Table \ref{tab:results_gdap}.

\begin{table}[]
\centering
\fontsize{6pt}{8pt}\selectfont
\resizebox{\columnwidth}{!}{%
\begin{tabular}{lllll}
\hline
\multicolumn{2}{l}{\textbf{Dataset}}       & \textbf{P} & \textbf{R} & \textbf{F-1} \\ \hline
\multirow{2}{*}{\textbf{MLEE}}       & All & 73.05      & 76.74      & 74.85        \\ 
                                     & Pos & 73.97      & 77.49      & 75.69        \\ \hline
\textbf{RAMS}                        & All & 72.61      & 71.62      & 72.11        \\ \hline
\multirow{2}{*}{\textbf{MAVEN}}      & All & 59.01      & 63.82      & 61.32        \\ 
                                     & Pos & 60.5       & 64.67      & 62.51        \\ \hline
\multirow{2}{*}{\textbf{WikiEvents}} & All & 61.29      & 63.33      & 62.29        \\ 
                                     & Pos & 56.73      & 57.61      & 57.17        \\ \hline
\end{tabular}%
}
\caption{Results using alternative evaluation scheme on all datasets. All: Multi-label metrics on all rows, with NONE as a separate class. Pos: Multi-label metrics on instances with at least one event. Multi-class and multi-word triggers count as distinct labels, with an exact match, counted as true positive.}
\label{tab:results_gdap}
\end{table}

\section{Annotation issues}
\label{app:annotation}


We present an approach that accurately extracts text terms for event annotations while preserving case sensitivity, a crucial factor in distinguishing different event triggers. 
Improper extraction or human error can lead to errors in existing annotations. Our approach can identify such errors by highlighting discrepancies in the case of event triggers. 
Additionally, we observe an ambiguity in some annotation schema, particularly in MAVEN, where the extensive coverage of event types results in overlapping event type definitions. 
For instance, the event types motion, self\_motion, and motion\_direction exhibit minor differences, leading to inconsistent annotations. 
This ambiguity introduces noise into the classification and ED subtasks. Our proposed model resolves this issue and accurately extracts all events in the corpus. 
We provide examples that demonstrate the improved ED performance achieved through multi-tasking.

\section{EDM3 improves single-task ED performance on WikiEvents}
\label{EDM3_vs_singletask}

\noindent
\fbox{
\begin{minipage}{\linewidth}
{ 
   \textbf{\newline Input:} \\ Police in Calais have dispersed a rowdy anti-migrant protest with tear gas after clashes with protesters and detained several far-right demonstrators. \\
   \newline
    \textbf{Single-task:} \\ detained->movement.transportperson \\ 
    \newline
    \textbf{EDM3:} \\ detained->movement.transportperson | \textbf{clashes\\->conflict.attack} \\ 
    \newline
    \textbf{Gold:} \\ detained->movement.transportperson | \textbf{clashes\\->conflict.attack} \\ 
}
\end{minipage}
}

\begin{figure*}
    \centering
    \includegraphics[width=14cm]{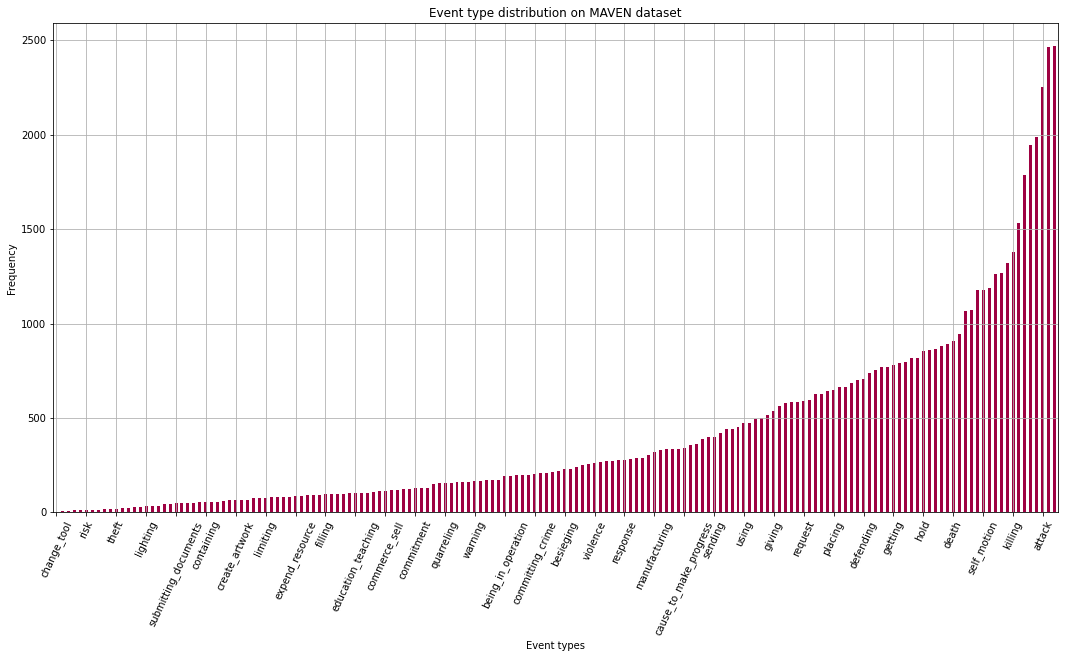}
    \caption{Distribution of event types in MAVEN}
    \label{fig:maven_dtb}
\end{figure*}

\begin{figure*}
    \centering
    \includegraphics[width=14cm]{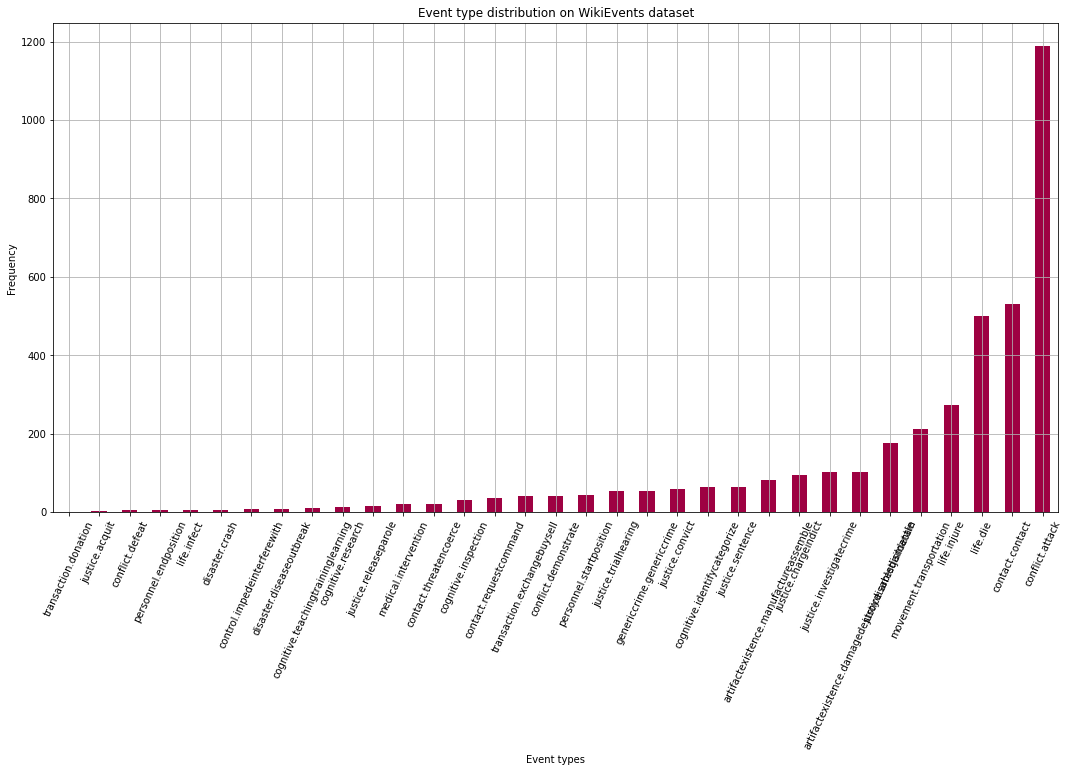}
    \caption{Distribution of event types in WikiEvents}
    \label{fig:wiki_dtb}
\end{figure*}

\begin{table}[]
\fontsize{12pt}{\baselineskip}\selectfont
\resizebox{\columnwidth}{!}{%
\begin{tabular}{|l|l|l|}
\hline
\textbf{Category}           & \textbf{Event type} & \textbf{Example triggers}           \\ \hline
\multirow{7}{*}{Anatomical} & cell\_proliferation & proliferation, proliferate, growing \\
        & development                & formation, progression, morphogenesis \\
        & blood\_vessel\_development & angiogenic, angiogenesis              \\
        & death                      & death, apoptosis, survival            \\
        & breakdown                  & dysfunction, disrupting, detachment   \\
        & remodeling                 & remodeling, reconstituted             \\
        & growth                     & proliferation, growth, regrowth       
        \\ \hline
\multirow{6}{*}{Molecular}  & synthesis           & production, formation, synthesized  \\
        & gene\_expression           & expression, expressed, formation      \\
        & transcription              & expression, transcription, mRNA       \\
        & catabolism                 & disruption, degradation, depleted     \\
        & phosphorylation            & phosphorylation                       \\
        & dephosphorylation          & dephosphorylation                     
        \\ \hline
\multirow{5}{*}{General}    & localization        & migration, metastasis, infiltrating \\
        & binding                    & interactions, bind, aggregation       \\
        & regulation                 & altered, targeting, contribute        \\
        & positive\_regulation       & up-regulation, enhancement, triggered \\
        & negative\_regulation       & inhibition, decrease, arrests         \\ \hline
Planned & planned\_process           & treatment, therapy, administration    \\ \hline
\end{tabular}%
}
\caption{Event types in MLEE, along with example triggers.}
\label{tab:mlee_types}
\end{table}
\begin{table}[t!]
\resizebox{\columnwidth}{!}{%
\begin{tabular}{|l|l|l|}
\hline
\textbf{Event type} & \textbf{Frequency} & \textbf{Example triggers}     \\ \hline
process\_start      & 2468          & began, debut, took place      \\
causation           & 2465          & resulted in, caused, prompted \\
attack              & 2255          & bombing, attacked, struck     \\
hostile\_encounter  & 1987          & fought, conflict, battle      \\
motion              & 1944          & fell, pushed, moved           \\
catastrophe         & 1785          & explosion, hurricane, flooded \\
competition         & 1534          & event, championships, match   \\
killing             & 1380          & killed, murder, massacre      \\
process\_end        & 1323          & closing, complete, ended      \\
statement           & 1269          & asserted, proclaimed, said   
\\ \hline
\end{tabular}%
}
\caption{Top 10 event types in MAVEN, along with example triggers.}
\label{tab:maven_types}
\end{table}
\begin{table}[t!]
\resizebox{\columnwidth}{!}{%
\begin{tabular}{|l|l|l|}
\hline
\textbf{Event type}           & \textbf{Frequency} & \textbf{Example triggers}        \\ \hline
conflict.attack               & 721           & massacre, battle, bombing        \\
movement.transportperson      & 491           & smuggling, walked, incarcerate   \\
transaction.transfermoney     & 482           & reimbursed, paid, purchasing     \\
life.die                      & 442           & die, murder, assassinating       \\
life.injure                   & 422           & surgery, injured, brutalized     \\
movement.transportartifact    & 367           & imported, trafficking, smuggling \\
transaction.transferownership & 327           & auction, donated, acquire        \\
contact.requestadvise         & 250           & advocating, recommending, urged  \\
contact.discussion            & 249           & discuss, meet, negotiated        \\
transaction.transaction       & 211           & funded, donated, seized
\\ \hline
\end{tabular}%
}
\caption{Top 10 event types in RAMS, along with example triggers.}
\label{tab:rams_types}
\end{table}

\begin{table*}[]
\resizebox{\textwidth}{!}{%
\begin{tabular}{|l|l|l|}
\hline
\textbf{Event type}                             & \textbf{Frequency} & \textbf{Example triggers}           \\ \hline
conflict.attack                                 & 1188               & explosion, shot, attack             \\
contact.contact                                 & 530                & met, said, been in touch            \\
life.die                                        & 501                & killed, died, shot                  \\
life.injure                                     & 273                & injuring, wounded, maimed           \\
movement.transportation                         & 212                & transferred, brought, arrived       \\
justice.arrestjaildetain                        & 176                & arrested, capture, caught           \\
artifactexistence.damagedestroydisabledismantle & 103                & damaged, destruction, removed       \\
justice.investigatecrime                        & 102                & analysis, discovered, investigation \\
justice.chargeindict                            & 96                 & charged, accused, alleged           \\
artifactexistence.manufactureassemble           & 82                 & construct, make, build   
\\ \hline
\end{tabular}%
}
\caption{Top 10 event types in WikiEvents, along with example triggers.}
\label{tab:wiki_types}

\end{table*}
\begin{table*}[]
\resizebox{\textwidth}{!}{%
\begin{tabular}{|
l |
l |
l |}
\hline
\textbf{WikiEvents} &
  \textbf{Common} &
  \textbf{RAMS} \\ \hline
\begin{tabular}[c]{@{}l@{}}conflict.defeat,\\  medical.intervention,\\  disaster.diseaseoutbreak,\\  justice.releaseparole,\\  movement.transportation,\\  cognitive.inspection,\\  justice.acquit,\\  justice.sentence,\\  transaction.exchangebuysell,\\  justice.trialhearing,\\  cognitive.identifycategorize,\\  justice.convict,\\  artifactexistence.damagedestroydisabledismantle,\\  genericcrime.genericcrime,\\  artifactexistence.manufactureassemble,\\  contact.requestcommand,\\  control.impedeinterferewith,\\  justice.investigatecrime,\\  justice.chargeindict,\\  cognitive.teachingtraininglearning,\\  transaction.donation,\\  cognitive.research,\\  life.infect,\\  disaster.crash,\\  contact.contact\end{tabular} &
  \begin{tabular}[c]{@{}l@{}}justice.arrestjaildetain,\\  personnel.startposition,\\  personnel.endposition,\\  conflict.attack,\\  conflict.demonstrate,\\  life.injure,\\  contact.threatencoerce,\\  life.die\end{tabular} &
  \begin{tabular}[c]{@{}l@{}}contact.collaborate,\\  justice.investigate,\\  contact.commitmentpromiseexpressintent,\\  justice.judicialconsequences,\\  contact.mediastatement,\\  contact.commandorder,\\  manufacture.artifact,\\  contact.negotiate,\\  transaction.transaction,\\  government.legislate,\\  contact.publicstatementinperson,\\  contact.funeralvigil,\\  disaster.fireexplosion,\\  artifactexistence.damagedestroy,\\  government.formation,\\  justice.initiatejudicialprocess,\\  government.agreements,\\  personnel.elect,\\  movement.transportperson,\\  transaction.transferownership,\\  conflict.yield,\\  inspection.sensoryobserve,\\  government.spy,\\  government.vote,\\  transaction.transfermoney,\\  movement.transportartifact,\\  disaster.accidentcrash,\\  contact.discussion,\\  contact.requestadvise,\\  contact.prevarication\end{tabular} \\ \hline
\end{tabular}%
}
\caption{Event types in RAMS and WikiEvents. Common: list of event types common to both datasets.}
\label{tab:rams_wiki_types}
\end{table*}

\end{document}